\documentclass[conference]{IEEEtran}
\usepackage{times}

\usepackage[numbers]{natbib}
\usepackage{multicol}
\usepackage[bookmarks=true]{hyperref}

\usepackage{amsmath}
\usepackage{amssymb}
\usepackage{hyperref}
\usepackage{cleveref}
\usepackage{color}
\usepackage[sc]{mathpazo}
\usepackage{cite}
\usepackage{graphicx}
\usepackage{makecell}
\usepackage{comment}

\newcommand{\figref}[1]{Fig. \ref{#1}}

\begin{document}

\title{On complementing end-to-end\\ human behavior predictors with planning}

\author{Liting Sun, Xiaogang Jia, and Anca D. Dragan\\ University of California, Berkeley}



%

\maketitle

\begin{abstract}
High capacity end-to-end approaches for human motion (behavior) prediction have the ability to represent subtle nuances in human behavior, but struggle with robustness to out of distribution inputs and tail events. Planning-based prediction, on the other hand, can reliably output decent-but-not-great predictions: it is much more stable in the face of distribution shift (as we verify in this work), but it has high inductive bias, missing important aspects that drive human decisions, and ignoring cognitive biases that make human behavior suboptimal. In this work, we analyze one family of approaches that strive to get the best of both worlds: use the end-to-end predictor on common cases, but do not rely on it for tail events / out-of-distribution inputs -- switch to the planning-based predictor there. We contribute an analysis of different approaches for detecting when to make this switch, using an autonomous driving domain. We find that promising approaches based on ensembling or generative modeling of the training distribution might not be reliable, but that there very simple methods which can perform surprisingly well -- including training a classifier to pick up on tell-tale issues in predicted trajectories. 
\end{abstract}

\IEEEpeerreviewmaketitle
\section{Introduction}
\label{sec:intro}
Robots that need to share their environments with humans learn predictive models of human behavior, which they use to generate their own behavior in response. Autonomous cars try to  predict where other cars will go \citep{driggs2018robust, schmerling2018multimodal} and what pedestrians will do \citep{ma2017forecasting}, indoor mobile robots try to predict where the people around them will move \citep{ziebart2009planning}, and manipulators try to predict how human collaborators will reach for objects in their workspace \citep{mainprice2013human,lasota2017multiple, ding2011human, koppula2013anticipating, lasota2015analyzing}.

When choosing the function class for these learned predictors, high capacity models are very appealing. Recent progress has shown that we can train deep neural networks end-to-end to go from a history of raw state information or even raw sensor data to a distribution over predicted trajectories for a human, implicitly or explicitly extracting relevant features, identifying potential targets in the scene, computing trajectories for each, and assessing their relative likelihoods \citep{chai2019multipath, liang2020learning, zhao2020tnt, zeng2021lanercnn}. Such models dominate the leaderboards in benchmarks for motion prediction (or "forecasting", as it is sometimes referred to) like Argoverse \citep{chang2019argoverse} or INTERACTION \citep{zhan2019interaction}. They free us from specifying what features might be important or identifying a "theory of mind" for how humans make decisions. Their capacity enables them to represent subtle nuances of human behavior, like people's implicit proxemics preferences, risk aversion level, or anything else that influences where humans go that would be otherwise very challenging to explicitly write down. 

But one challenge that such high capacity, end-to-end models face is their performance in the face of distribution shift or tail events. Our understanding of the nuances of this challenge is still evolving, but there seem to be at least two phenomena at play: one stemming from the model's capacity itself, and one stemming from the way we train these models. 

On the capacity side, the function class can represent so many hypotheses that there will exist many of them which fit the data, some based on spurious correlations rather than on the underlying human decision making process that generated those motions. The learner will not be able to disambiguate among them, and can converge to a hypothesis based on a correlate. Over-parametrization increases the ability to represent such hypotheses, lowering average error but possibly increasing error for tail events \citep{sagawa2020investigation}.

\begin{figure}
    \centering
    \includegraphics[width=\columnwidth]{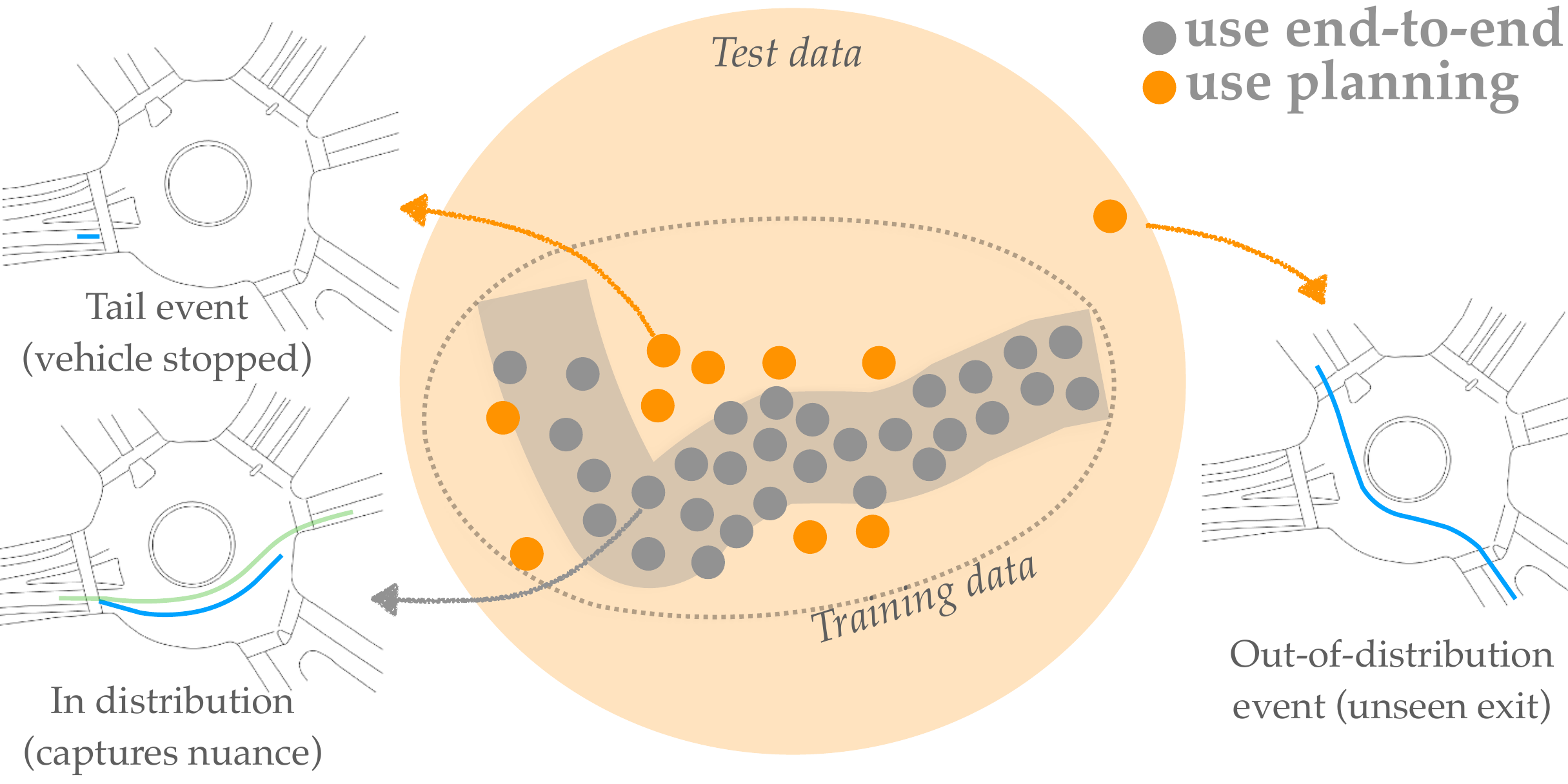}
    \caption{We analyze methods for using an end-to-end predictor on common cases (gray region), and relying on planning-based prediction outside of that (orange region). }
    \label{fig:front}
\end{figure}


Then on the training side, stochastic gradient descent methods introduce their own biases. Sometimes, this bias helps with the capacity issue by pushing the optimization away from the correlates and enabling generalization \citep{soudry2018implicit}. However, if there is an easy correlate in the data that helps get most examples right, training might "lazily" converge to that instead of identifying the (more complex) causal variables that explain the data. For instance, a model might learn to predict braking only once the brake light is on, instead of solving the more complex problem of identifying the need for human to begin braking. This is the more problematic the more information the model has access to, because more correlates exist (the "causal misindentification" \citep{de2019causal}.

On the other hand, we have \emph{planning}-based predictors. These are based on the idea that human motion results from decisions people make in pursuit of their goals and preferences. Learn their goals and preferences, and leverage planning to generate the corresponding motions \citep{ziebart2009planning, sadigh2016planning, kretzschmar2016socially}. Sticking to the braking example: this type of predictor will easily figure out that a human will need to brake without needing to see their brake lights first. It learns that people want to make progress but avoid collisions and stop at stop signs, and it uses a planner to generate a motion that will do just that. If slowing down is necessary for collision avoidance or because a stop sign is coming up, that is what the planner will do. And indeed, prior work has shown such predictors to be preferable in highly interactive domains, depending on how one collects their training data \citep{choudhury2019utility}. 

However, planning-based predictors suffer from too much inductive bias. They commit to predefined notions (features) for what humans care about which are inevitably both inaccurate (e.g. humans do care about collision avoidance, but might be more sensitive to front collisions than side collisions, for instance), as well as missing important aspects altogether \citep{fisac2018probabilistically}. Further, real people are far from optimal decision makers: we have all sorts of systematic biases, from perception biases to wrong beliefs to risk-aversion to optimism bias and beyond. In short, these predictors are not expressive enough to capture the nuances in human behavior. 

Naturally, we would want the best of both worlds. In this work, we analyze one way to strive for that: \emph{use the end-to-end predictor on common cases, but do not rely on it for rare / out-of-distribution inputs -- switch to the planning-based predictor there} (\figref{fig:front}). On these inputs, even though the planning-based predictor will not be able to perfectly anticipate human behavior, it will still get the basics right (in our driving domain, for instance, it will output trajectories that stay on the road, avoid collisions, etc.).

We contribute an analysis of different approaches for detecting when to make this switch. We start by outlining natural ways to solve the problem, from detecting out-of-distribution inputs by ensembling and generative methods, to learning to classify when the predictor failed, to using real-time observations of the human's current motion to detect that the predictor is doing a poor job at anticipating what is happening. In order to assess and compare their ability to make more accurate predictions on difficult inputs by switching to the planning-based predictor when they have to, but keeping the end-to-end one when it performs well, we create tests sets that purposefully introduce domain shift. We measure each method's ability to accurately identify the shift, as well the resulting "hybrid" predictor's accuracy. 

Our findings first support this hypothesized relationship between end-to-end predictors and planning-based ones. Our end-to-end predictor does much better in-distribution (on a validation set drawn from the same distribution as the training data) than the planning-based one. On the other hand, it is not robust to the shifts and perturbations we introduce, whereas the planning-based one stays remarkably consistent. One of our contributions is merely showcasing this in the driving domain.

As for the switching methods, the results are quite interesting. Training an ensemble and using disagreement as a stand in for "the predictor is uncertain here" \citep{lakshminarayanan2016simple} fails to identify many of the cases that it should, because the members agree even when making the wrong prediction. Switching based on observing the real human online to take actions that the predictor is assigning low probability to is very reliable, but it does introduce a significant switching delay because the robot has to observe enough such "low probability" human actions. Surprisingly, a simple classifier that we train to label predictions as good or bad based on the predictor's performance on training data is also very reliable. The classifier is not meant to quantify uncertainty or detect out of distribution issues, but it implicitly does that by learning to pick up on features of predicted trajectories that are suggestive of something having gone wrong, like going off the road or not stopping at a stop sign.

Overall, if we can combine the power of end-to-end prediction with the robustness we get out of planning-based prediction, robots that act around people will be able to anticipate and adapt to nuanced human behavior while still maintaining reasonable performance in the long tail. Our work analyzes one family of approaches that strive for this combination, with somewhat surprising but promising results. We intend this as a first step in this direction, starting a discussion into what approaches and ideas are most promising when it comes to this general goal of getting the best out of both worlds.

\section{Methods}
\label{sec:method}

\subsection{Problem Statement}
We are given a training set $\mathcal{D}$ consisting of tuples  $(x, h,\xi)$, where $x$ is the state of the environment (including map data), $h$ is the history of motion observation for all agents in the scene, and $\xi$ is the trajectory label for the target agent. We are also given two predictors trained on (a subset of) this data: 1) a high capacity "end-to-end" model $f_{e2e}$ that learns to map $(x,h)$ to $\xi$; 2) a planning-based model $f_{plan}$ that learns a cost function that explains the motions observed in training, and optimizes it to generate predictions $\xi$ for the target agent (and the other agents in the scene).\footnote{Note that planning-based models need to make explicit joint or iterative predictions about other agents as well, since the target agent optimizes to avoid collisions with them.}

The goal is to output a switching detector $\sigma: (x,h,f_{e2e}(x,h))\mapsto \{0,1\}$ that determines, based on a new input $(x,h)$ from a test distribution $\mathcal{T}$ and optimally the prediction $\hat{\xi}=f_{e2e}(x,h)$ on that new input, whether this is an input on which the end-to-end predictor will have high error. Needless to say, $\sigma$ does not get access to the test distribution $\mathcal{T}$.

Armed with $\sigma$, the robot, upon encountering a new input $(x,h)$, can predict an agent's trajectory using 
$$f_{\sigma}(x,h)=\begin{cases}f_{e2e}(x,h), & \sigma(x,h,f_{e2e}(x,h))=0 \\ f_{plan}(x,h), & \sigma(x,h,f_{e2e}(x,h))=1\end{cases}$$
We discuss below four natural ways to train such a $\sigma$.

Aside: Note that this makes the assumption that the end-to-end model will only fail in rare or out-of-distribution situations, where our best bet will be the planning model. Depending on how these are trained, this will not always be true (end to end models might break on common cases as well, and planning-based predictors might not be better on rare cases). Nonetheless, we choose to focus on this setting because we believe it to be most representative of how these predictors will be developed in the real world, where end to end models will have high enough capacity to fit the average case well, and planning-based models will be of low enough capacity to generalize well. Our experiments below support these assumptions.

\subsection{Preliminaries: Example Predictors}
We use the INTERACTION dataset \citep{zhan2019interaction} to train our predictors. We use segments of the data of 40 timesteps from the beginning, middle, and end of runs (10 timesteps as the history, and the other 30 timesteps as the label). We train a LSTM model for our end-to-end predictor with pooling layers to encourage the interactions among agents in a similar way as social-LSTM \citep{alahi2016social} (see Appendix B for details), and we use Inverse Reinforcement Learning (IRL) to recover a cost function for our planning-based predictor using features for collision avoidance, progress, lane keeping, etc., similar to \citep{kuderer2015learning, sun2018probabilistic}. To enable the model to predict collision avoidance with other agents, we use the cost function to first optimize predictions for further away agents, and iteratively compute trajectories for nearby agents until we reach the target agent. To avoid counfounding effects from the need to predict the geometric intent of agents, we provide both predictors with the overall reference path the agent is following (e.g. which exit they are taking in a roundabout). Further details are in Appendix A. 

\subsection{Ensemble Disagreement}
Following \citep{lakshminarayanan2016simple}, we note that if we train an ensemble instead of a single end-to-end predictor, ensemble disagreement can be used as a measure of uncertainty or confidence. When the members of the ensemble make contradicting predictions, this is a signal that we are out-of-distribution.

We select 5 different models as the ensemble members. All the members share the same structure as the end-to-end predictor, and they all train with the same training data. However, each of them is trained with different initialization, and the order of examples they use naturally differ because of SGD (stochastic gradient descent). To quantify their disagreement, we take the most-probable prediction from each of them, denoted by $\hat{\xi}_j (j=1,2,\cdots,5)$, and calculate the variance on their final positions. Namely, suppose the final positions on $\hat{\xi}_j$ is $(x^F_j, y^F_j)$, then the metric for the disagreement is given by\footnote{We experimented with several distance metrics in our experiment and this performed the best, though the results are very sensitive to the choice of a metric, so we encourage practitioners to both analyze different metrics when trying this, as well as different ways of creating diversity in the ensemble -- ours is but a starting point, so we used the simplest approach.}
$$E_{disagree} = \max\{var\{x^F_{j=1,\cdots,5}\}, var\{y^F_{j=1,\cdots,5}\}\}$$
$$\sigma(x,h)=\begin{cases}1, &E_{disagree}\geq \tau \\ 0, &otherwise \end{cases}$$
with $\tau$ a threshold we tune on the training set.

\subsection{Generative Modeling of the Training Distribution}
Another way to detect whether a new input is in the training distribution is to train a generative model of the inputs -- train a GAN (generative adversarial network) on inputs $(x,h)$, and use its discriminator at test time as our $\sigma$ to tell whether the test input is "real", i.e. from the training data, or not.

\begin{figure}[h]
    \centering
    \includegraphics[width=\linewidth]{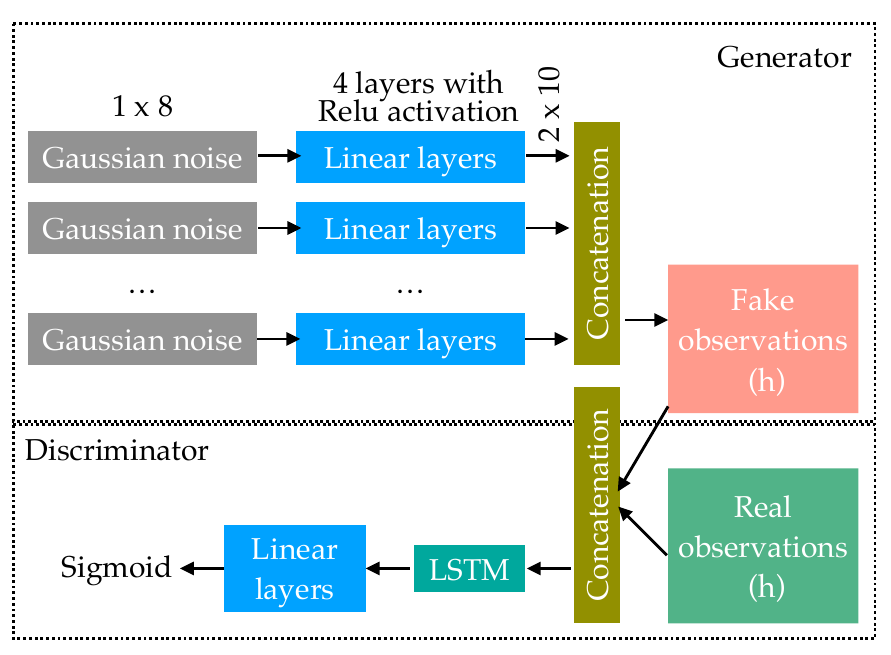}
    \caption{The structure for the generative model}
    \label{fig:GAN_model}
\end{figure}

We use the architecture from \figref{fig:GAN_model}. In our experiments we only focus on the $h$ side of the input (the history) and not the whole scene $x$ because of the difficulty in generating scene configurations, but in principle that would be possible as well. The generated data is labeled as "fake", the training data is labeled as "real".

\subsection{Classifying Poor Predictions}
The third approach is to simply train a classifier to explicitly distinguish whether a prediction is good enough. That is, instead of attempting to detect whether an input is out of distribution or rare, simply look at the prediction and classify whether it's correct -- with the hope that discriminating that a prediction is poor is an easier task for a neural network that producing a good prediction in the first place. The classifier tries to learn a function $\sigma:(x,h,f_{e2e}(x,h)) \mapsto \{0, 1\}$ from training data $\mathcal{D}_{\text{classifier}}$ that we auto-label based on the average distance error (ADE) between the predicted trajectory and the ground-truth trajectory in the predictor training data. We label all predicted trajectories that generate large ADEs (2 sigmas beyond the mean ADE of the training set) as bad predictions (1).

\begin{figure}[h]
    \centering
    \includegraphics[width=\linewidth]{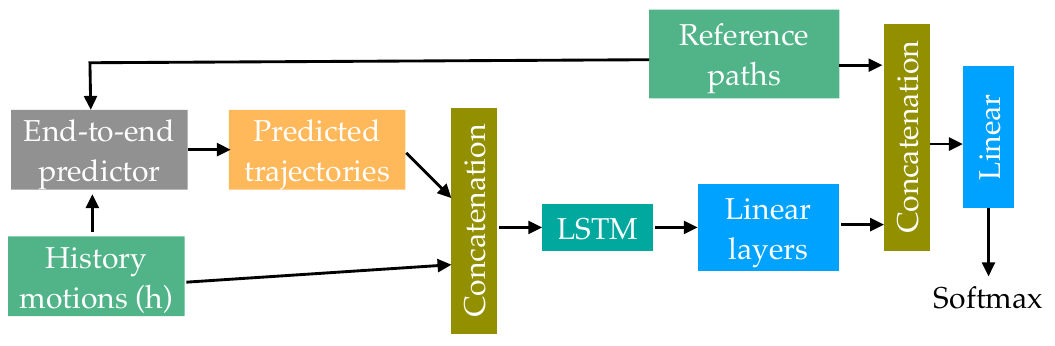}
    \caption{The structure for the classifier}
    \label{fig:classifier_model}
\end{figure}

\figref{fig:classifier_model} shows the structure of the classifier, similar to that in the discriminator of the generative modelling approach (but with access to the predicted trajectory as well). We use a softmax loss for training. 

\subsection{Online (Bayesian) Failure Estimation}
The fourth approach we investigate is a bit different in what it has access to. Rather than training something ahead of time, here the robot at test time receives ground truth human observations from the human it is trying to predict, and uses them to determine whether its predictor is operating accurately or not in that particular new setting. This is directly inspired by \citep{fisac2018probabilistically}: if at test time the human takes actions that are too low of a probability under the predictor, we conclude that the predictor is not correctly handling the current situation. 

\begin{figure}[h]
    \centering
    \includegraphics[width=\linewidth]{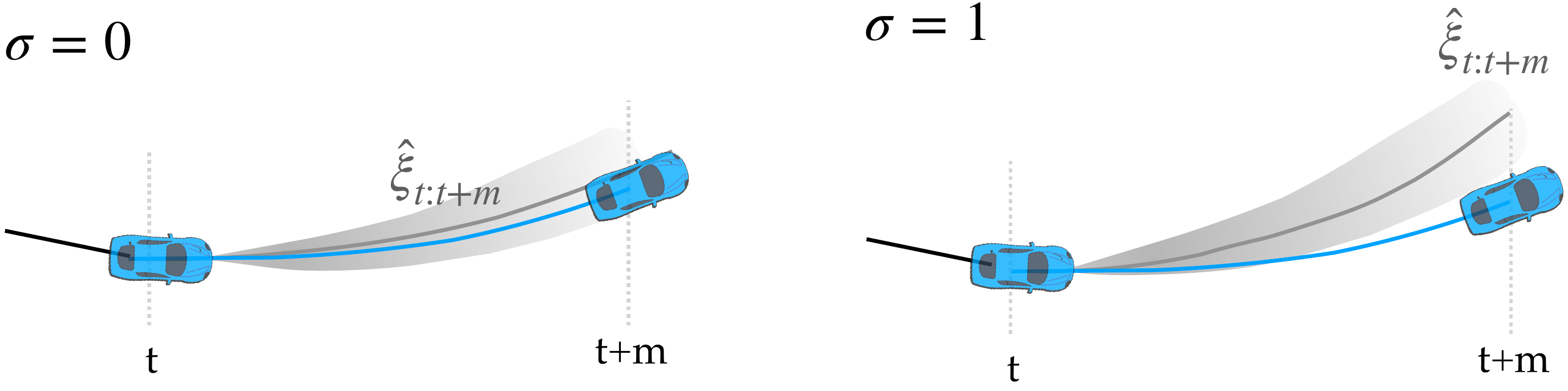}
    \caption{An illustrative example of the Bayesian failure estimation approach: gray bands are two different predicted distributions at time $t$ and the blue one is the new observed trajectory at time $t+m$.}
    \label{fig:bayesian}
\end{figure}

Suppose that at time $t$, we have the predicted trajectory as $\hat{\xi}_t = f_{e2e}(x, h)$. After another $m$ timesteps ($m$ is smaller than the length of the predicted trajectory), we have new human observations, denoted as $\xi_{t:t+m}$. From the discrepancy between $\xi_{t:t+m}$ and $\hat{\xi}_t$, we can infer whether the prediction $\hat{\xi}_t$ is good or not. An illustrative example is given in \figref{fig:bayesian}.

In general, for probabilistic predictors where $f_{e2e}(x, h)$ is a probability distribution, we can use
$$\sigma(\xi_{t:t+m}, f_{e2e}(x, h))=\begin{cases}1, &P(\xi_{t:t+m}|f_{e2e}(x, h))<\tau \\ 0, & otherwise\end{cases}$$
with $\tau$ a threshold tuned based on the training data. For predictors that only output trajectories, we can use a proxy distribution based on distance, $P(\xi | \hat{\xi}_t)\propto \exp\{-\dfrac{1}{m}L_2(\hat{\xi}_{t:t+m},\xi_{t:t+m})\}$.

\section{Experiments}
\label{sec:experiments}

To analyze how promising these switching approaches are for detecting that we should plug in the planning-based predictor instead of using the default end-to-end one, we need data where tail events and distribution shift occur. While tail events will happen for most training distributions we would encounter, distribution shift is something we decided to purposefully introduce in a controlled way. We use a real driving data set (the INTERACTION data \citep{zhan2019interaction}), and design three experiments that use different train and test sets to probe at different types of shift -- from introducing noise to the input, to testing on a new exit from a roundabout where we didn't have data (new reference paths), all the way to switching to an entirely new map. 

Note these shifts sound more drastic than they are. We chose them not because autonomous cars might navigate new maps -- since the predictors only focus on 30-step snippets of the overall trajectory and have access to the geometric reference that the vehicle is following, this is more akin to changing the configurations of the road and other agents than putting the robot in an entirely new region. But as experiment designers, it frees us from having to slice the data by what are common configurations vs. new ones -- if we had a metric for this, that would be our switching method in the first place!

\subsection{Experiment Design}
\noindent\textbf{Independent Variables.} We manipulate which switching method $\sigma$ is used, from our four methods in \cref{sec:method}, and adding always 0 (use $f_{e2e}$ only) and always 1 (use $f_{plan}$ only). We also manipulate whether shift is present or not, and the type of shift: new reference paths (new exit), new map, and noise. 

\begin{figure}[t!]
	\begin{centering}
		\includegraphics[width=\linewidth]{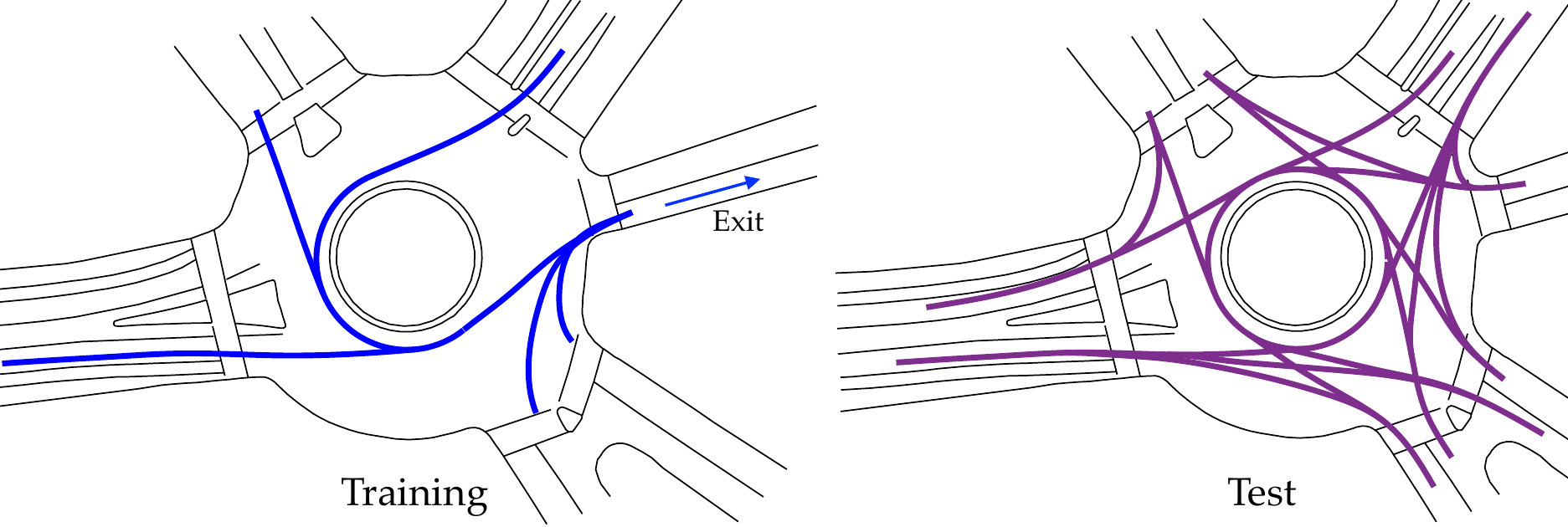}
	\end{centering}
	\caption{The training and test domains for experiment I, where we introduce new exits. The predictors only deal with local 30-step snippets and have access to the ground truth reference path.}
	\label{fig:exp1_settings}
\end{figure}

\begin{figure}[t!]
\centering
		\includegraphics[width=0.35\linewidth]{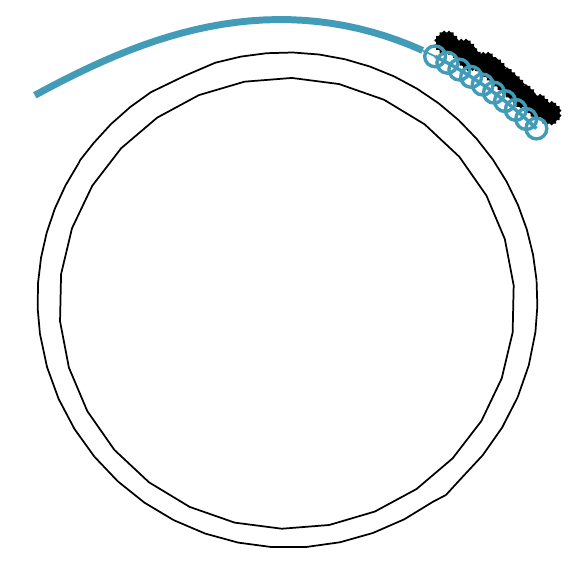}
	\caption{The training and testing domain  for experiment III, where we introduce noise (blue circle: real observations; black star: observations with added noise; blue curve: ground truth future trajectories).}
	\label{fig:exp3_settings}
\end{figure}

In Experiment I, as shown in \figref{fig:exp1_settings}, all the training data includes trajectories that exit only through one selected lane (left), but in the test domain, trajectories following all possible reference paths are included (right). In the training domain, 1948 examples were included.

In Experiment II, we train on the map for Experiment I (all exits, 39764 examples), but test on different maps (still providing the reference as an input). 

In Experiment III, we train on all maps, and add Gaussian noise to the history of observations at test time. We set the Gaussian parameters as $\mu=0.5, \sigma=0.1$, and the effect this has is visualized in \figref{fig:exp3_settings}.

The switching methods have access to the same training data as the predictors. For each experiment, we also have a validation set from the same distribution on the training data, and evaluate the methods on both validation and test.  

\noindent\textbf{Metrics.} 
We have two measures: accuracy and performance. Accuracy refers to the switching method's ability to predict whether the end-to-end predictor will have high error. Performance refers to the resulting hybrid predictor (which uses planning-based when $\sigma=1$) error, measured as average distance error (ADE) between the predicted trajectories and the ground-truth ones, which is standard in motion prediction \citep{alahi2016social, chai2019multipath, zhao2020tnt, zeng2021lanercnn, liang2020learning}. 

\noindent\textbf{Hypotheses.} We hypothesize that H1) $f_{e2e}$ has lower validation set ADE than $f_{plan}$, but higher test set ADE; and H2) $f_{\sigma}$ (the hybrid predictor) has lower ADE than both in both validation and test by using $f_{plan}$ on tail events and out of distribution inputs where $f_{e2e}$ struggles. However, the goal of the analysis is to compare the different approaches for $\sigma$ and establish their potential strengths and weaknesses.

\subsection{Experiment I - Generalizability across different references}

\begin{figure}[ht!]
	\begin{centering}
		\includegraphics[width=\linewidth]{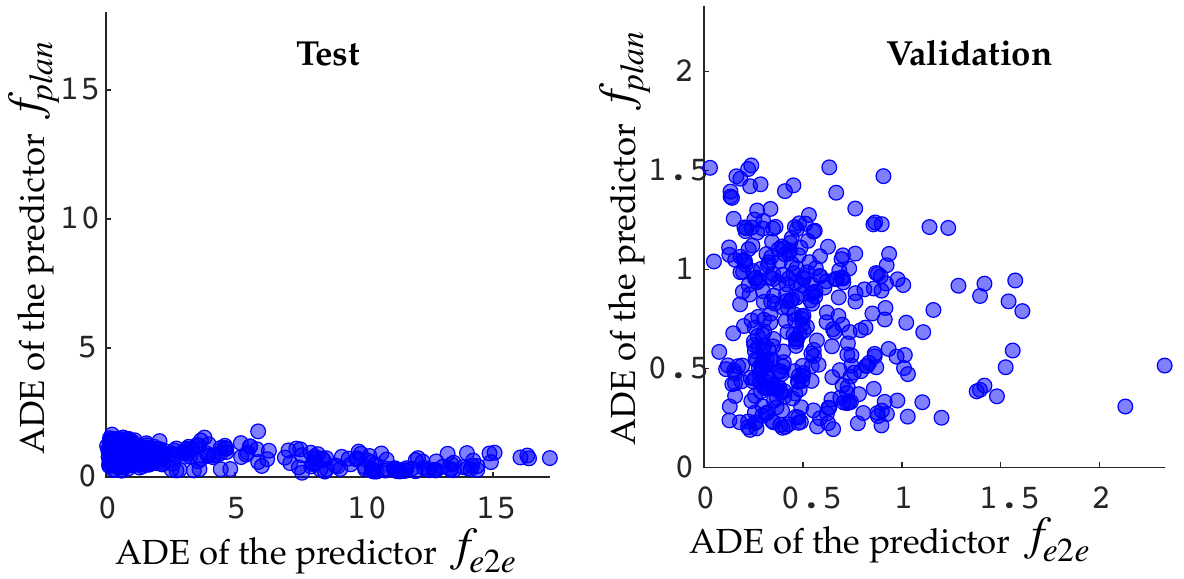}
	\end{centering}
	\caption{The performance of the two predictors in validation and test domains in Experiment I. Note the difference in scale.}
	\label{fig:exp1_scatter}
\end{figure}

\noindent\textbf{Performance of the predictors.}
\figref{fig:exp1_scatter} shows the performance of the two predictors in the validation and test sets. In validation, $f_{e2e}$ (LSTM) has lower error than $f_{plan}$ (IRL) -- .527 vs .721 (see Table \ref{tab:results_exp_1}). There are some tail events with high error though. On test, the opposite is true. This is in line with H1. Note that $f_{plan}$ has relatively steady performance in the face of the shift, whereas $f_{e2e}$ goes from much better to drastically worse.

\begin{table}[t!]
	\centering
	\begin{tabular}{|p{3cm}|p{0.95cm}|p{0.95cm}|p{0.95cm}|p{0.95cm}|}
		\hline
		&  \makecell{Val acc.}& \makecell{Test acc.} & \makecell{Val \\ADE (m)} & \makecell{Test \\ADE (m)} \\
		\hline
		$f_{e2e}$ only (LSTM) & & & 0.527 & 2.9648 \\
		\hline
		$f_{plan}$ only (IRL) & && 0.721 & 0.80 \\ 
		\hline
		ensemble & 81.78\% & 83.05\% & 0.5464 & 1.2699\\ 
		\hline
         GAN  & 75.13\% & 84.53\% & 0.5191 & 0.9118 \\ 
		\hline
		classifier & {\bf 85}\% & 85\% & 0.5377 & {\bf 0.6804} \\ 
		\hline
		30-step online Bayesian  &{\bf 100}\% & {\bf 100}\%& {\bf 0.4177} & {\bf 0.6162}\\ 
		\hline
		5-step online Bayesian  &68.3\% & {\bf 89}\%& {\bf 0.4892} & 0.8235\\ 
		\hline
	\end{tabular}
	\caption{Results of experiment I}
	\label{tab:results_exp_1}
\end{table}

\begin{figure}[ht!]
	\begin{centering}
	\includegraphics[width=0.95\linewidth]{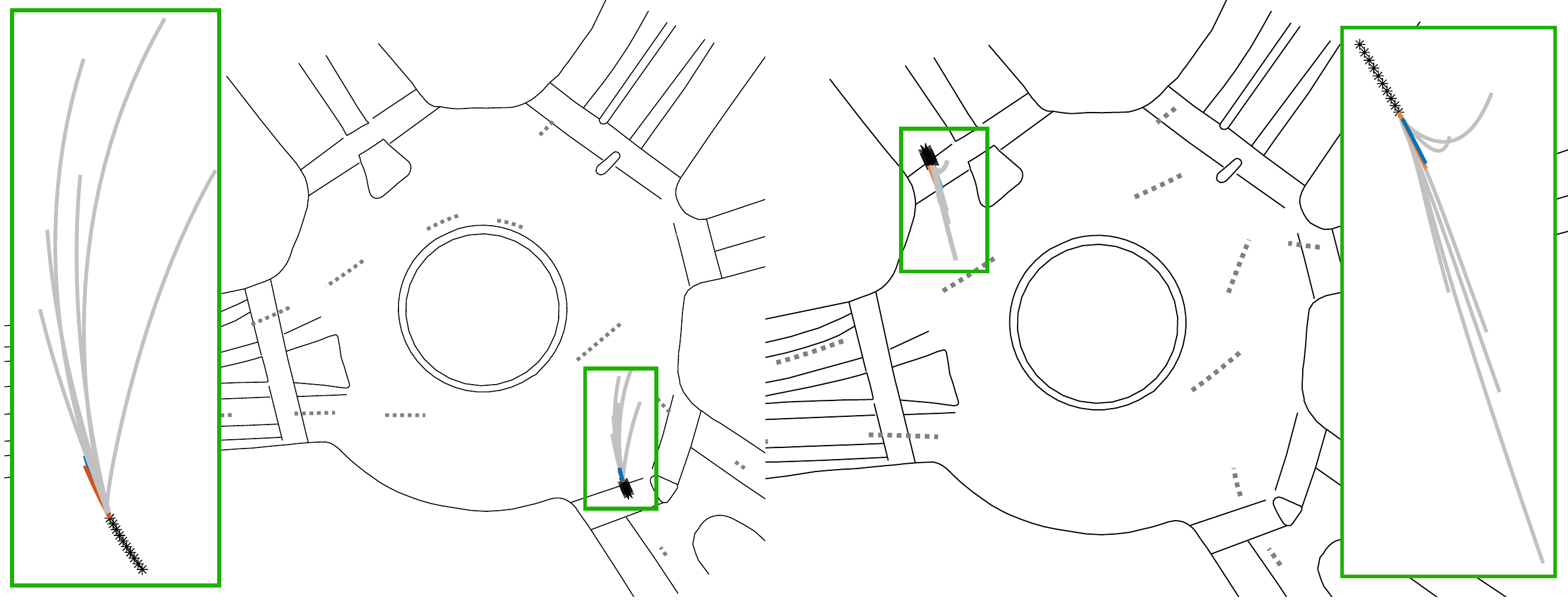}
	\end{centering}
	\caption{A failure mode of $f_{e2e}$ on the training domain, where the predictions (gray) make progress when the ground truth (blue, see the zoomed-in details) almost stays put. The black dotted lines in the center are observations for surrounding vehicles.}
	\label{fig:exp1_failure_lstm_val}
\end{figure}

\noindent\textbf{Failure modes of the predictors.} 
\emph{Training domain: }Many of the $f_{e2e}$ failures were in cases where the vehicle is stopped or moving slowly due to stop signs or traffic congestion. These are relatively rare events. \figref{fig:exp1_failure_lstm_val} shows two such cases where the predictor assumes progress when the ground truth stays put.

\begin{figure}[ht!]
	\begin{centering}
	\includegraphics[width=0.95\linewidth]{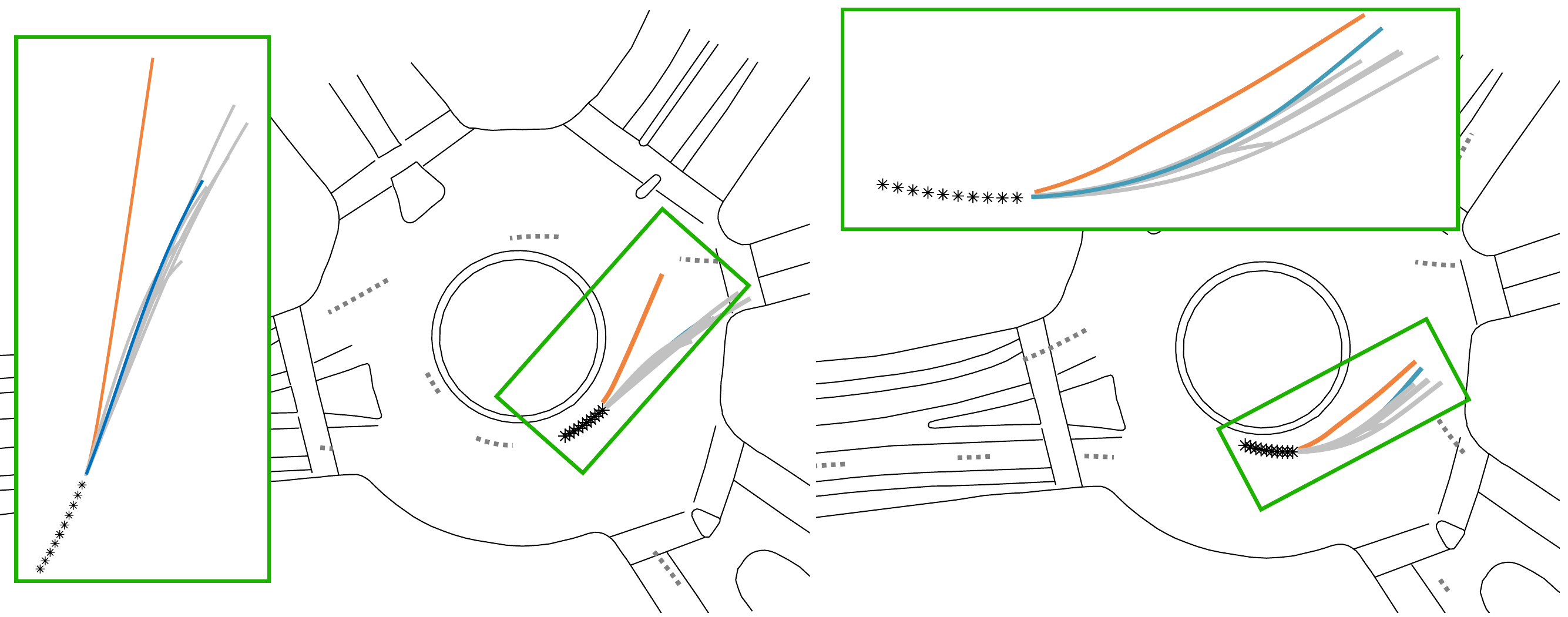}
	\end{centering}
	\caption{Failure modes of $f_{plan}$ on the training domain, where it has the wrong reference path (left) or missing features / converging to poor optima (right). The $f_{plan}$ predictions are in orange, $f_{e2e}$ in gray, ground truth in blue (similar to $f_{plan}$).}
	\label{fig:exp1_failure_irl_val}
\end{figure}

$f_{plan}$ has no issue predicting the stop because it models people as following rules and avoiding collisions. However, it struggles on cases where its structure is wrong. Unlike $f_{e2e}$, it uses the reference path as a hard constraint, and some of our data has the wrong reference (\figref{fig:exp1_failure_irl_val}, left). This is akin to what might happen in the real world where the map annotations are wrong, and a planning-based predictor will stick to them. It also sometimes creates unnatural motion (\figref{fig:exp1_failure_irl_val}, right) because it is missing important aspects of what people want or converging to bad local optima.

\begin{figure}[ht!]
	\begin{centering}
	\includegraphics[width=0.9\linewidth]{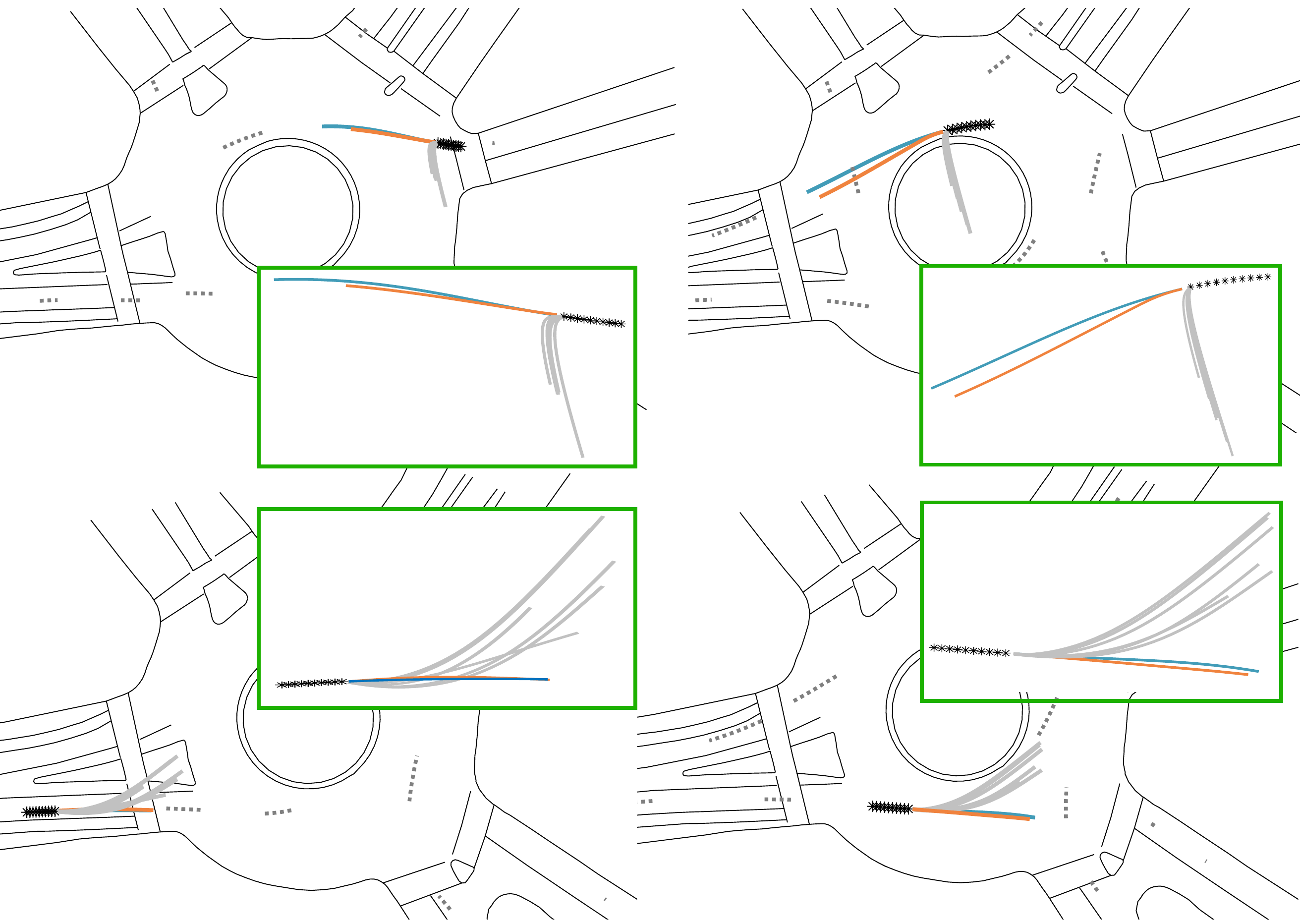}
	\end{centering}
	\caption{Failures of $f_{e2e}$ in the test domain for Experiment I. The predictions in gray aggressively pursue the wrong exit and/or ignore road obstacles. $f_{plan}$ predictions are in orange, similar to ground truth (blue).}
	\label{fig:exp1_failure_lstm_test}
\end{figure}

\emph{Testing domain: }
$f_{e2e}$ fails at test time either by failing to pursue the correct reference (e.g. predicting drastic turns to go to the exit it was trained on), or by ignoring road geometry and moving through obstacles. Both are likely to be due to the shift in distribution. \figref{fig:exp1_failure_lstm_test} shows some examples.
	
\begin{figure}
	\centering
	\includegraphics[width=0.95\linewidth]{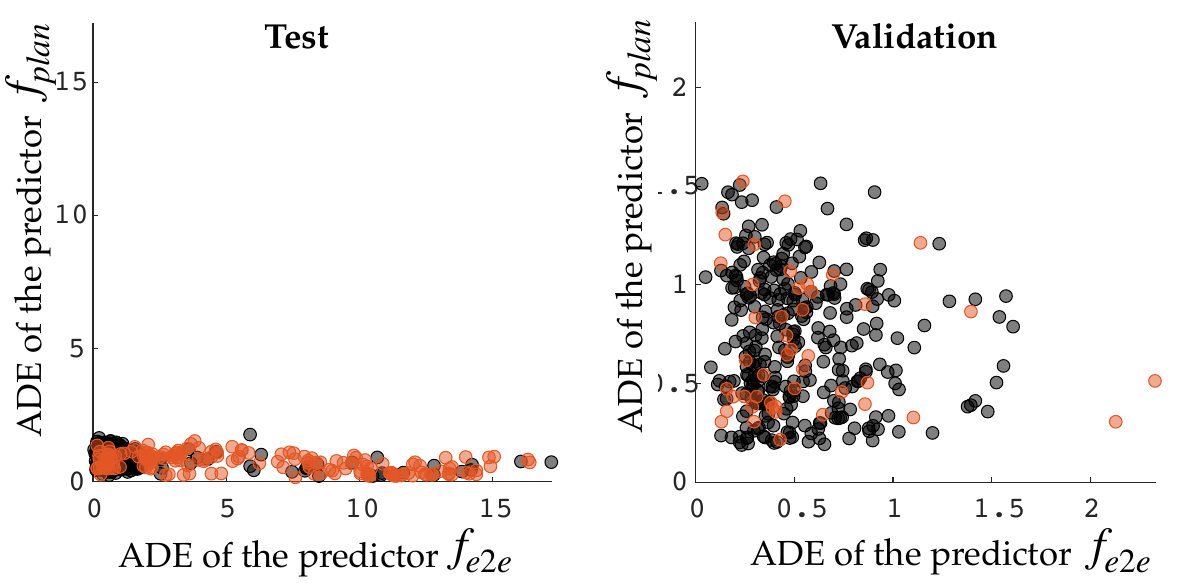}
	\caption{The ADE scatter with ensembling as a performance detector (black: good cases for the lstm model; red: good cases for the irl model)}
	\label{fig:scatterwithensemble}
\end{figure}

\begin{figure}
	\centering
	\includegraphics[width=0.95\linewidth]{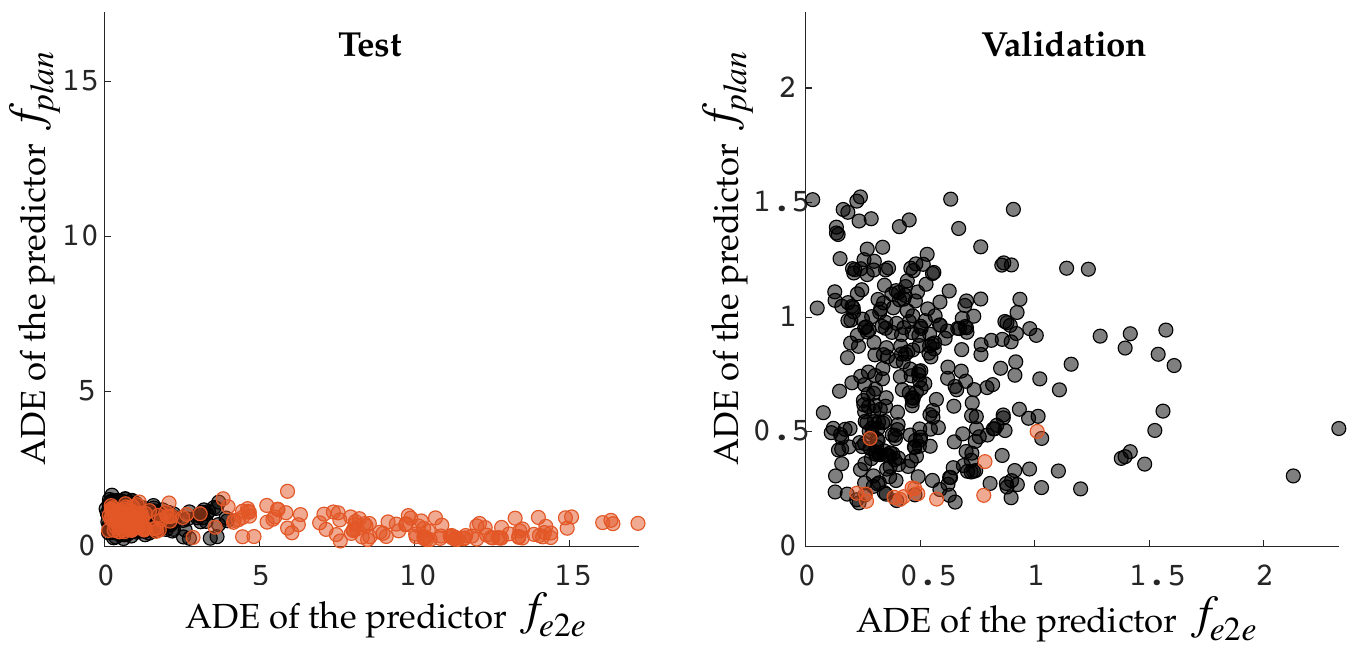}
	\caption{The ADE scatter with GAN as a scenario detector (black: good cases for the lstm model; red: good cases for the irl model)}
	\label{fig:scatterwithgan}
\end{figure}

\begin{figure}
	\centering
	\includegraphics[width=0.95\linewidth]{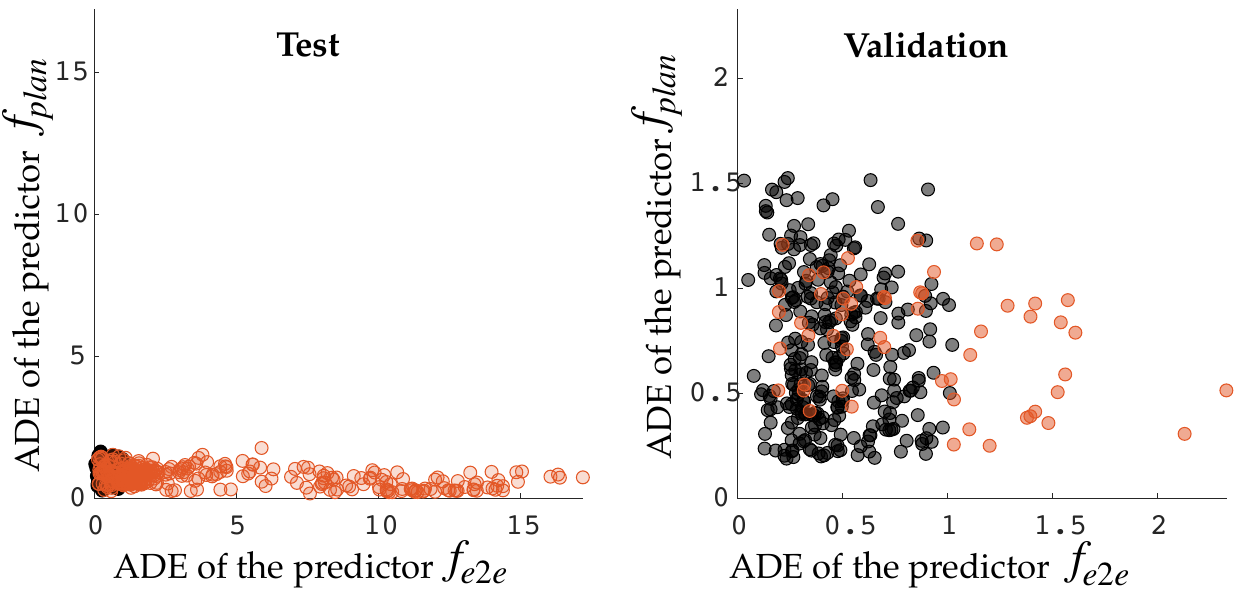}
	\caption{The ADE scatter with a trained classification model as a performance detector (black: good cases for the lstm model; red: good cases for the irl model)}
	\label{fig:scatterwithclassification}
\end{figure}

\begin{figure}
	\centering
	\includegraphics[width=0.95\linewidth]{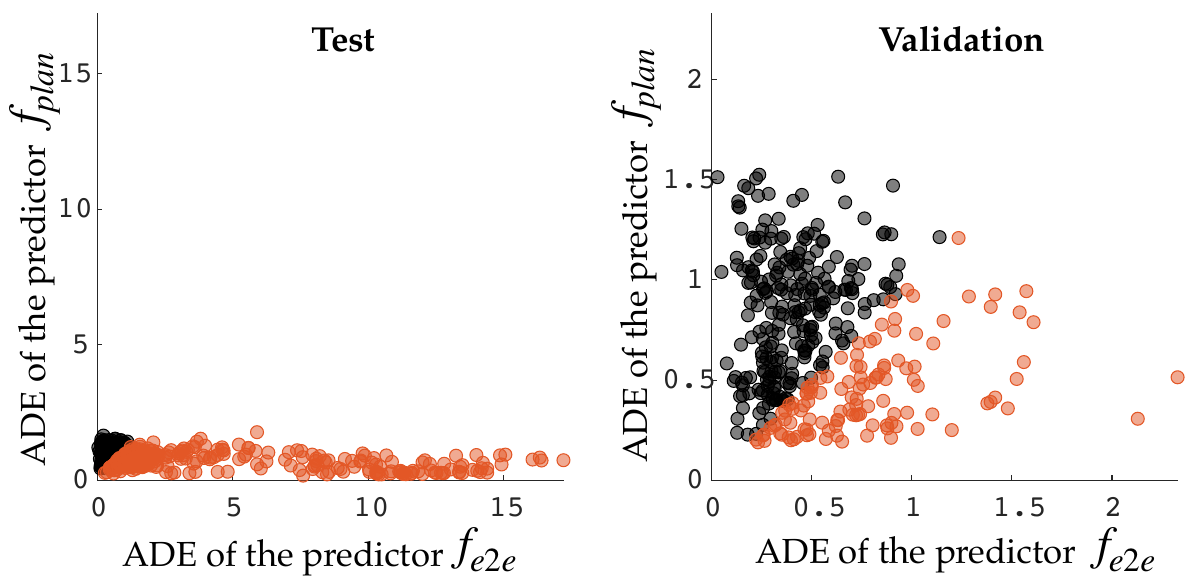}
	\caption{The ADE scatter with online bayesian based on 30-step trajectories (black: good cases for the lstm model; red: good cases for the irl model)}
	\label{fig:scatterwithfulltrajbayesian}
\end{figure}

\noindent\textbf{Performance of switching approaches.} \figref{fig:scatterwithensemble} - \figref{fig:scatterwithfulltrajbayesian} show the four approaches ability to pick up on high error for $f_{e2e}$ in validation and test. Right off the bat, we see that the ensemble struggles -- sometimes it agrees on data it shouldn't (high error on x axis), and disagrees on data it should agree on (low error on x axis). It does catch the biggest outliers in validation, but fails to catch some high error points in test. This could be explained by the bias in SGD -- despite the fact that there are many hypotheses explaining the data that the class can represent, and ideally the ensemble members would converge to different ones, in reality the bias in SGD itself might lead to converging to similar hypotheses. 

The GAN has mixed performance, though better. Many of the test cases look similar enough to the training data that the discriminator misses them.

On the other hand, the classifier works remarkably well (\figref{fig:scatterwithclassification}, picking up on many issues in validation and test, with some false positives. This is interesting, because the classifier is only trained in-domain, so how can it give the correct label off-distribution? The answer is that there are enough failures in the training domain (on those tail events) that the classifier learns the "tell-tale signals" that the prediction has gone wrong. Instead of focusing on the input domain itself (like the GAN), it focuses on the prediction, and picks on patterns like "whenever the prediction goes off the road, we get high error", or "whenever there is a stop sign and the prediction is not stopping, we get high error". While this will not catch more subtle wrong predictions, it seems to be an effecting way of automatically learning metrics for sanity checking predictions (without having to think ahead of time of what these metrics need to be and craft them by hand).

\begin{figure}
	\centering
	\includegraphics[width=0.9\linewidth]{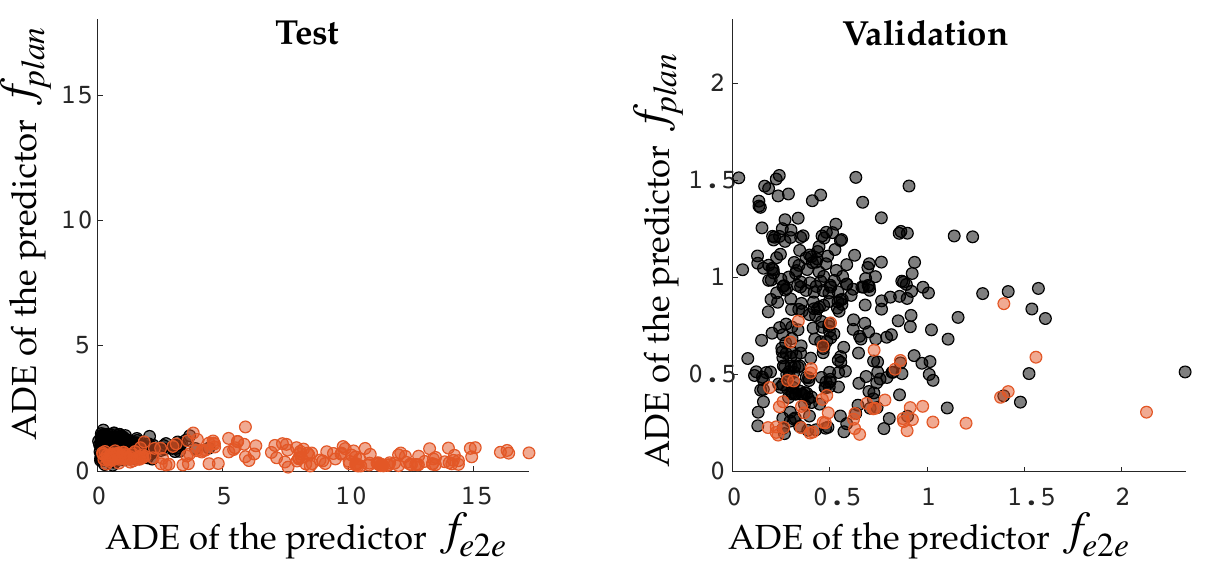}
	\caption{The ADE scatter with online bayesian based on 5-step trajectories (black: good cases for the lstm model; red: good cases for the irl model)}
	\label{fig:scatterwith5steptrajbayesian}
\end{figure}

When given enough time steps of observation (\figref{fig:scatterwithfulltrajbayesian}), the online failure detector works perfectly. Unfortunately, this is not very practical because it causes a long delay between the predictions being poor and switching to $f_{plan}$. When we try to make the call with only 5 steps (\figref{fig:scatterwith5steptrajbayesian}), results are much worse. 

Table \ref{tab:results_exp_1} shows what effects this accuracy end up having for the $f_{\sigma}$ error. On validation, all methods maintain the good performance of $f_{e2e}$, with the online failure detector improving it. We see here a discrepancy between accuracy and error, where even if the 5-step online detector has lower accuracy than other methods, it ends up better error. This is perhaps because the yes/no metric does not capture the magnitude of the error.  On the test set, all methods drastically improve performance, with the classifier and online failure detector performing the best.  



\subsection{Experiment II - Generalizability across different maps}
\noindent\textbf{Performance of the predictors.} The performance of the two predictors matches Exp I: $f_{plan}$ is stable, $f_{e2e}$ is much better in validation and much worse in test, as shown in Table \ref{tab:results_exp_2}, as well as in \figref{fig:exp2_scatter} in the Appendix C.

\begin{table}[t!]
	\centering
	\begin{tabular}{|p{3cm}|p{0.95cm}|p{0.95cm}|p{0.95cm}|p{0.95cm}|}
		\hline
		&  \makecell{Val acc.}& \makecell{Test acc.} & \makecell{Val \\ADE (m)} & \makecell{Test \\ADE (m)} \\
		\hline
		$f_{e2e}$ only (LSTM) & && 0.367 & 1.691 \\
		\hline
		$f_{plan}$ only (IRL) & && 0.8224 & 1.0634 \\ 
		\hline
		ensemble &80.3\% & 82.4\%& 0.3888 & 0.9369\\ 
		\hline
		classifier  & {\bf 93}\%&{\bf 93}\%& 0.372 & {\bf 0.742} \\ 
		\hline
		30-step online Bayesian  &{\bf 100}\% & {\bf 100}& {\bf 0.3280} & {\bf 0.6435}\\ 
		\hline
		5-step online Bayesian  &52.3\% & 85\%& {\bf 0.3475} & 0.8461\\ 
		\hline
	\end{tabular}
	\caption{Results of experiment II}
	\label{tab:results_exp_2}
\end{table}

\begin{figure}[ht!]
	\begin{centering}
 	\includegraphics[width=\linewidth]{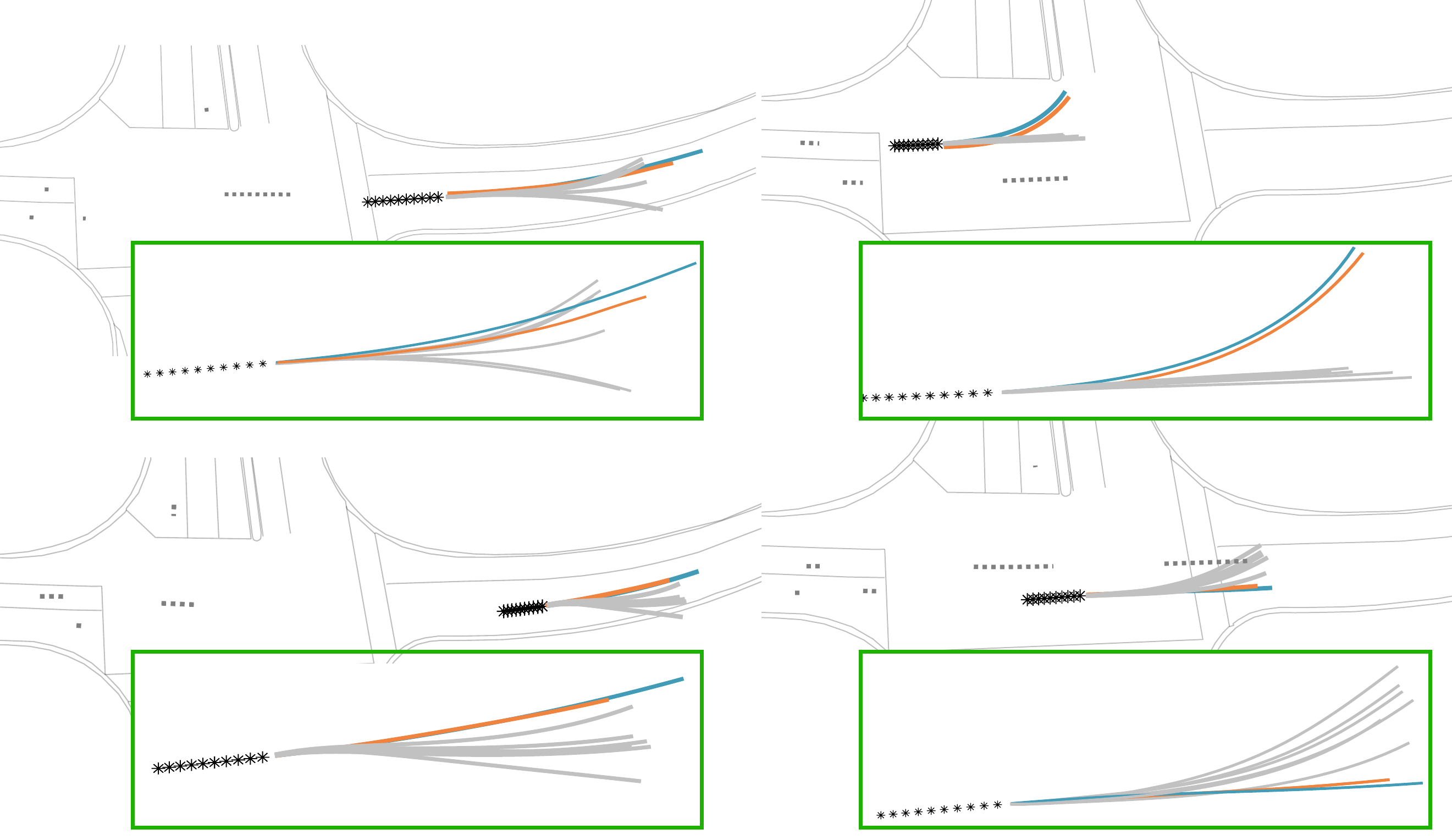}
	\end{centering}
	\caption{Failures of $f_{e2e}$ when tested on a new map. Its predictions in gray, $f_{plan}$ predictions in orange, and ground truth in blue.}
	\label{fig:exp2_failure_lstm_test}
\end{figure}

\noindent\textbf{Failure modes of the predictors.}
We see that with this testing domain, $f_{e2e}$ sometimes fails to follow the reference, or produces trajectories that stray off the road (\figref{fig:exp2_failure_lstm_test}), while $f_{plan}$'s structure enables it to seamlessly produce reasonable trajectories despite the difference in domain. The Appendix C provides further examples in \figref{fig:exp2_failure_lstm_test_2} from another test map. 

\noindent\textbf{Performance of switching approaches.}
Table \ref{tab:results_exp_2} summarizes the findings, mirroring Exp I. 

\subsection{Experiment III - Robustness to added noise}

\noindent\textbf{Performance of the predictors.}
The performance of the two predictors is analogous, although a bit more extreme than in the previous two experiments (see Table \ref{tab:results_exp_3}): here, the end-to-end predictor really struggles under Gaussian noise so its test domain error is drastically higher. $f_{plan}$ remains relatively stable, but takes a bit of a hit as well compared previous settings.  

\begin{table}[t!]
    \centering
    \begin{tabular}{|p{3cm}|p{0.95cm}|p{0.95cm}|p{0.95cm}|p{0.95cm}|}
		\hline
		&  \makecell{Val acc.}& \makecell{Test acc.} & \makecell{Val \\ADE (m)} & \makecell{Test \\ADE (m)} \\
         \hline
         $f_{e2e}$ only (LSTM) &&& 0.3561 & 7.0858 \\
         \hline
         $f_{plan}$ only (IRL) &&& 0.8194 & {\bf 1.0434} \\
         \hline
         ensemble & 79.24\% & 82\% & 0.3900 & 2.2693\\ 
         \hline
         GAN hybrid & 71.68\% & 87.44\% & 0.3561 & 1.4649 \\ 
         \hline
         classifier & {\bf 92.8}\% & {\bf 99.21}\% & 0.349 & 1.0588 \\
         \hline
         30-step online Bayesian & {\bf 100}\% & {\bf 100}\% & {\bf 0.2946} & {\bf 0.9016}\\ 
         \hline
         5-step online Bayesian & 75.34\% & 94.61\% & {\bf 0.3192} & 1.5211\\ 
         \hline
    \end{tabular}
    \caption{Results of experiment III}
    \label{tab:results_exp_3}
\end{table}

\begin{figure}[ht!]
	\begin{centering}
	\includegraphics[width=0.9\linewidth]{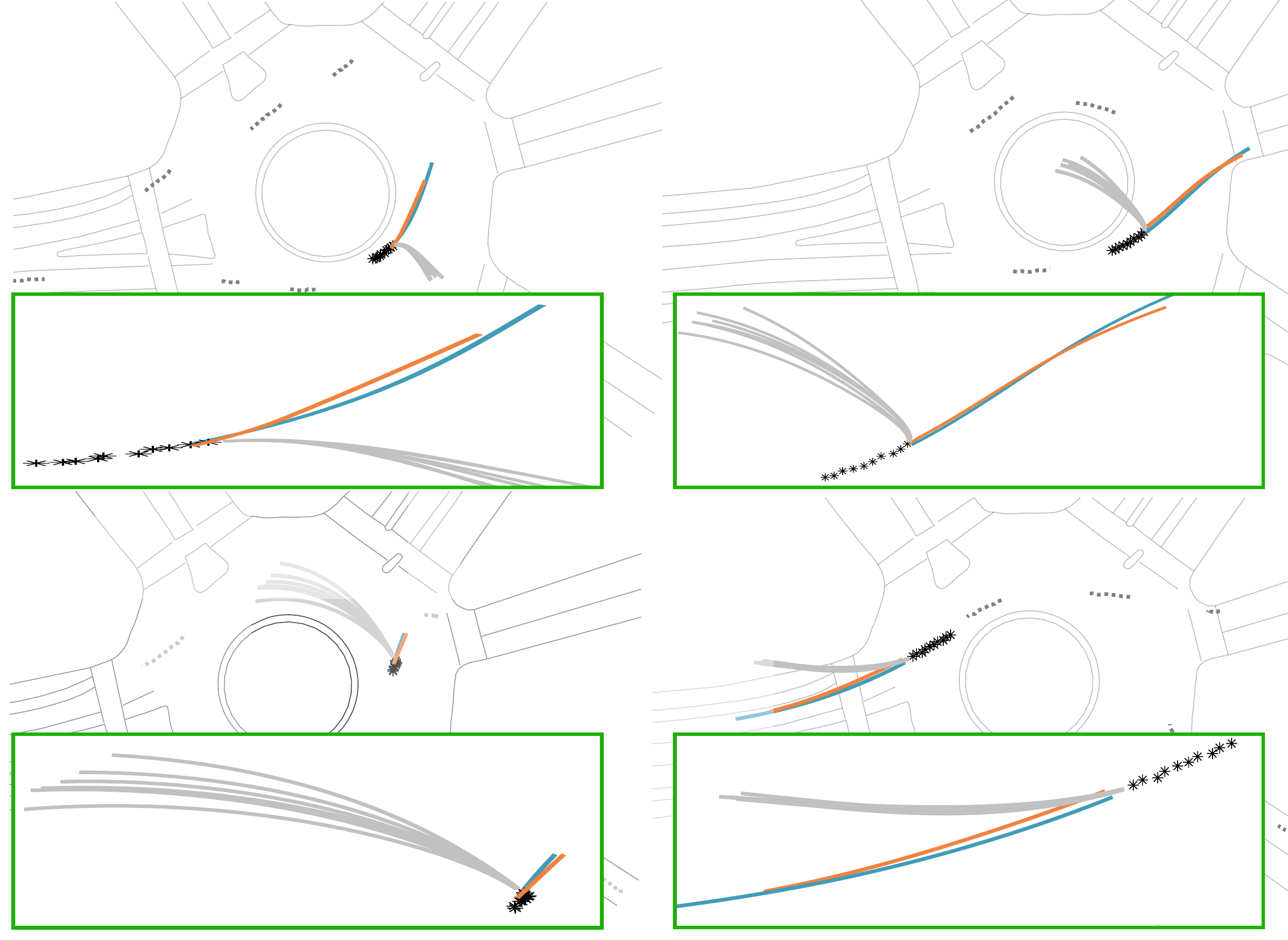}
	\end{centering}
	\caption{Failures of $f_{e2e}$ due to noise in the input (predictions in gray, compared to ground truth in blue and $f_{plan}$ in orange).}
	\label{fig:exp3_failure_lstm_test}
\end{figure}

\begin{figure}[ht!]
\centering
		\includegraphics[width=0.6\linewidth]{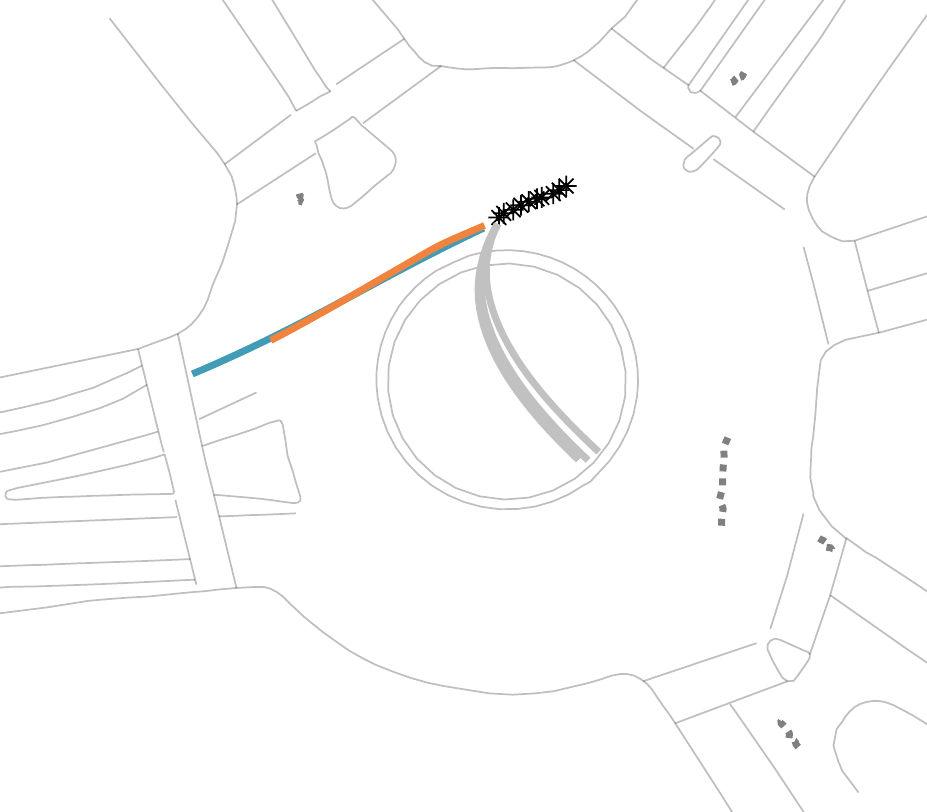}
	\caption{A failure of $f_{plan}$ due to noise in the input (predictions in orange, ground truth in blue speeds up more).}
	\label{fig:exp3_failure_irl_test}
\end{figure}

\noindent\textbf{Failure modes of the predictors.}
\figref{fig:exp3_failure_lstm_test} shows some failure cases due to the added noise for the $f_{e2e}$ predictor (qualitatively similar to the failures we saw in the previous experiments -- not following the reference and/or going off the road). $f_{plan}$ is also affected by the noise via wrong speed and orientation estimation for the vehicle, leading to less drastic errors but e.g. going much slower than the ground truth -- see \figref{fig:exp3_failure_irl_test}.

\noindent\textbf{Performance of switching approaches.}
Table \ref{tab:results_exp_3} shows the accuracy and resulting error for the different $\sigma$ approaches. Again, these reproduce what we see before in Experiments I and II. The Appendix C contains the analogous scatter plots for the error. 


\section{Discussion}
\label{sec:discussion}

\noindent\textbf{Summary of idea and approaches.} We put forward the hypothesis that high capacity end-to-end predictors will struggle with tail events and distribution shift, while planning-based predictors will be more robust but will not perform as well on common cases. We therefore want to get the best of both worlds, and we investigate one way to attempt this: detect when we're in one of those non-common cases, and switch to the planning-based method. We study natural approaches for this switch: 1) detecting that we are in a new area because multiple hypotheses that were consistent with training are disagreeing ("ensemble"), 2) discriminating that we're not in the training distribution by training a generative model of our data ("GAN"), 3) classifying whether the end-to-end model has high error by looking at the prediction it is outputting ("classifier"), and 4) detecting online that the model keep assigning low probability to what the current human is actually doing ("online Bayesian failure detector"). 

\noindent\textbf{Summary of findings.} Our first empirical contribution is to test our hypothesis above. By purposefully inducing distribution shift, we show just how robust planning-based predictors can be. This is important in and of itself, because in the era of end-to-end high capacity models we tend to forget the strengths of these high inductive bias methods -- it is intuitive that this robustness property would hold, but we quantify it in 3 experiments, and find in 2 of them a quite remarkable stability despite strong shift in the distribution. To the best of our knowledge, such a comparison between end-to-end models and planning-based approaches had not been performed.

Our second empirical contribution is a first analysis of these switching methods, which gives us some intuition about what to expect as they get further developed. We see that ensembles, while promising in theory, might not by default disagree when they should, perhaps because of the bias in SGD-like optimizers to converge to somewhat similar hypotheses despite stochasticity in initialization and sampling data. Also, we found that the ensembling performance relies on the metrics that we use. In contrast, we see that a classifier based on the predictor's performance in training data, which fires when the predictor is wrong, might actually be more powerful than it sounds: while it does not build a notion of being "out of distribution", it can learn to pick up on and recognize eggregious prediction mistakes (like going off-road). Of course, predictions that are subtly wrong (i.e. plausible) could escape such a method. It also seems like an online detector that figures out the predictor is mis-behaving and keeps attributing low probability to the human's actual actions is by far the most reliable -- of course, this would come at a delay, so it should be seen as a must-have fallback switching mechanism (our findings suggest everyone should use it, but also have other ways to detect a switch is needed that have less delay). 

\noindent\textbf{Limitations and future work.} Our work is limited in many ways. First, we do not study the prediction's impact on the robot's planning or behavior generation -- just improving prediction accuracy does not necessarily lead to robot behavior improvements (unless the robot itself works via imitation learning and we think of this method in that domain instead of the human modeling domain). Second, we studied very basic predictors (for both end to end as well as planning) and eliminated the need for intent prediction. Note however that the goal of our work is not to improve upon the latest state of the art predictors -- the point of our paper is that one can always improve the end to end learners with better architectures, data augmentation, better training, but the issues of domain shift and tail events are unlikely to go away. Finally, we operated under the hypothesis that we want to use the end-to-end predictor primarily (and indeed, in our results it superior to the planning-based approach for the common cases), but switching techniques based on treating the two predictors as experts and obtaining a mixture are also possible.  

Going further in this research, we are excited to pursue the switch idea as a data augmentation mechanism: switch to the planning-based predictor, but use the data generated this way to augment the real training data and therefore improve the robustness of the end-to-end model. 

\section*{Acknowledgments}

We thank the members of the InterACT lab for helpful discussion and ideas, especially Micah Carroll and Rohin Shah whose "fill-in-the-blanks" work was inspiration for this project. This work was partially supported by NSF CAREER and AFOSR.

\bibliographystyle{plainnat}
\bibliography{references}

\section{Appendix}
\label{sec:Appendix}
\subsection{The planning-based predictor}
\subsubsection{Learning the cost functions}
We assume that the cost function of human drivers is a linear combination of a set of predefined features. Thus, given a tuple $(x, h, \xi)$ in the training set $D$, the cost function associated with it can be described as $C({\xi}, \hat{\xi}_O; \mathbf{\pmb\theta})=\mathbf{\pmb\theta}^T\mathbf{f}({\xi}, \hat{\xi}_O)$. Note that $\hat{\xi}_O$ represents the estimated trajectories of all other surrounding agents, and $\mathbf{f}$ is the feature vector and $\mathbf{\pmb\theta}$ represents driver's preference over different elements in $\mathbf{f}$. With that, based on the principle of maximum entropy, we have
\begin{equation}
\label{eq:irl_maximum_entropy}
P({\xi} |\hat{\xi}_O; \mathbf{\pmb\theta}) \propto \exp\{-\beta C({\xi}, \hat{\xi}_O; \mathbf{\pmb\theta})\},
\end{equation}
where $\beta$ is a hyper-parameter that controls to what levels the human behaves as a rational optimizer.
Hence, the log-likelihood of the training set $D$ (with $N$ tuples) can be given by
\begin{equation}
\label{eq:irl_formulation}
\log P(D|\mathbf{\pmb\theta}) = \sum_{i=1}^N\log\dfrac{\exp\{-\beta C({\xi}_i, \hat{\xi}_{O, i}; \mathbf{\pmb\theta})\}}{\int \exp\{-\beta C(\Tilde{\xi}, \hat{\xi}_O; \mathbf{\pmb\theta})\}d\Tilde{\xi}}.
\end{equation}
By maximizing the log-likelihood, we can find the optimal parameter $\mathbf{\pmb\theta}^*$ that represents humans' preferences in real driving. 

\subsubsection{The feature set}
The features we selected to parametrize the trajectories in the planner-based predictors can be grouped as follows:
\begin{itemize}
\item Speed - The incentive of the human driver to reach a certain speed limit $v_{\lim}$ is captured by the feature
\begin{equation}
\label{eq:velocity_feature}
f_v({\xi}) = \sum_{t=0}^L (v_t-v_{\lim})^2
\end{equation}
$v_t$ is the speed at time $t$ along trajectory $\hat{\xi}$ and $L$ is the length of the trajectory in time.
\item Traffic - In dense traffic environment, human drivers tend to follow the traffic. Hence, we introduce a feature based on the intelligent driver model (IDM) \citep{kesting2010enhanced}
\begin{equation}
\label{eq:traffice_feature}
f_{\text{IDM}}({\xi}) = \sum_{t=0}^L (s_t-s_{t}^{\text{IDM}})^2
\end{equation}
where $s_t$ is the actual spatial headway between the front vehicle and predicted vehicle at time $t$ along trajectory $\hat{\xi}$, and $s_{t}^{\text{IDM}}$ is the spatial headway suggested by the IDM.
\item Control effort and smoothness - Human drivers typically prefer to drive efficiently and smoothly, avoiding unnecessary accelerations and jerks. To address such preference, we introduce a set of kinematics-related features:
\begin{equation}
\label{eq:kinematics}
f_{\text{acc}}({\xi}) =\sum_{t=0}^L a_t^2, \quad f_{\text{jerk}}(\hat{\xi}_M) =\sum_{t=1}^L \left(\dfrac{a_t-a_{t-1}}{\triangle t} \right)^2
\end{equation}
where $a_t$ represents the acceleration at time $t$ along the trajectory $\hat{\xi}_M$. $\triangle t$ is the sampling time.
\item Safety - Human drivers care about their distances to other agents to assure safety. Hence, we introduce a distance-related feature
\begin{equation}
\label{eq:distance_feature}
f_{\text{dist}}({\xi}) = \sum_{t=0}^L\sum_{k=1}^{n{+}1} e^{-\frac{(x_t-x^k_t)^2}{l^2}-\frac{(y_t-y^k_t)^2}{w^2}}
\end{equation}
where $(x_t, y_t)$ and $(x^k_t, y^k_t)$ represent, respectively, the coordinates of the target agent along $\hat{\xi}$  and those of the $k$-th surrounding vehicle. Parameters $l$ and $w$ are the length and width of the target agent. We use coordinates in Frenet Frame to be compatible with curvy roads, i.e., $x$ denotes the travelled distance along the road and $y$ is the lateral deviation from the target agent's lane center.
\item Goal - This feature describes the short-term goals of human drivers. Typically, goals are determined by the map and traffic control elements. To be more specific, we first check if there is any stop sign in the future horizon if the vehicle is driving at the speed limit. If there is stop signs, we will assign the short-term goal as the stop sign location. If not, we will set the short-term goal as the final position along the speed limit.
\begin{equation}
\label{eq:goal_feature}
f_{g}({\xi})= \sum_{t=0}^L\Vert(x_t, y_t)-(x_t^g, y_t^g)\Vert^2_2
\end{equation}
\end{itemize}
Note that we assume the geometric reference path is available for the target agent. Hence, Frenet Frame is utilized to integrate such information. Therefore, the predicted trajectories will always follow the reference path in the planning-based predictor.

\subsubsection{Iterative Predictions}
Note that in \eqref{eq:irl_formulation}, we need to estimate the future trajectories of other surrounding agents, i.e., $\hat{\xi}_{O}$. To achieve this, we used the planning-based predictor in an iterative way. More specifically, for each predicted agent, we consider all other surrounding agents if they are within a distance range with the predicted agents. Then we predict the future trajectories of agents in an reverse order based on their distances to the target agent. The farthest one is assumed to run at its current speed. Vehicles with lead-lag relationships are assumed to follow IDM, and others will be predicted based on the learned cost functions.

\subsection{The end-to-end predictor}
\begin{figure}[h!]
    \centering
    \includegraphics[width=\linewidth]{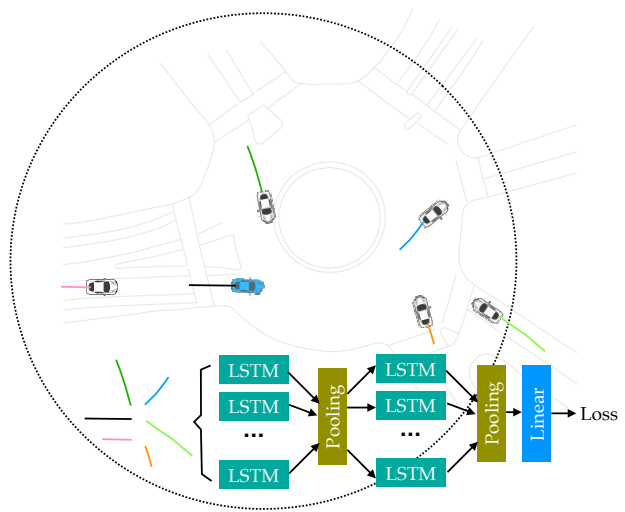}
    \caption{The structure of the lstm end-to-end predictor}
    \label{fig:lstm_model}
\end{figure}

The  end-to-end predictor we used in this work is backboned by the social LSTM model \citep{alahi2016social}. Similar to the planning-based predictor, we only consider surrounding agents which are within a distance range of the target agent. All agents' coordinates are centered based on the current (the most recent timestep in the history trajectory) position of the target vehicle. The historical motions of each agent is encoded separately with a long short-term memory (LSTM) module, and pooling layers on the hidden states are performed to encourage interactions among agents, as shown in \cref{fig:lstm_model}.

To capture the stochastic property of the trajectories, we optimize for the negative log-likelihood of the target's trajectories in the training set. More specifically, we assume that at each time step $t$, the position of the target agent can be represented as a Gaussian distribution parameterized by $(\mu_x^t, \mu_y^t, \sigma_x^t, \sigma_y^t)$. Hence, the linear module in the lstm predictor outputs $4\times L$ parameters, and the loss function is given by $\text{Loss}=-\sum_{\xi \in D}\sum_{t=1}^L\log P(x_t, y_t|\mu_x^t, \mu_y^t, \sigma_x^t, \sigma_y^t)$.

\subsection{Further Results / Details}
\label{sec:appendix_results}

\subsubsection{Experiment II}
\begin{figure}[ht!]
	\begin{centering}
		\includegraphics[width=\linewidth]{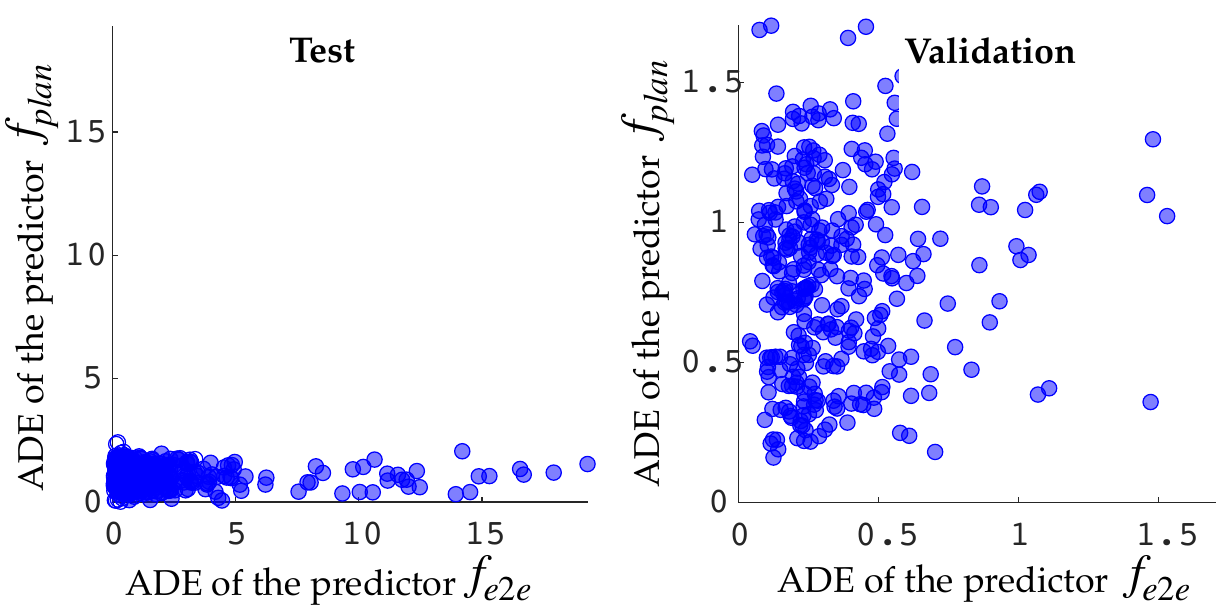}
	\end{centering}
	\caption{The performance of the two predictors in validation and test domains in Experiment II.}
	\label{fig:exp2_scatter}
\end{figure}

\figref{fig:exp2_scatter} shows the test and validation performance for the two predictors in Experiment II, similar to \figref{fig:exp1_scatter}.

\begin{figure}
	\centering
	\includegraphics[width=0.9\linewidth]{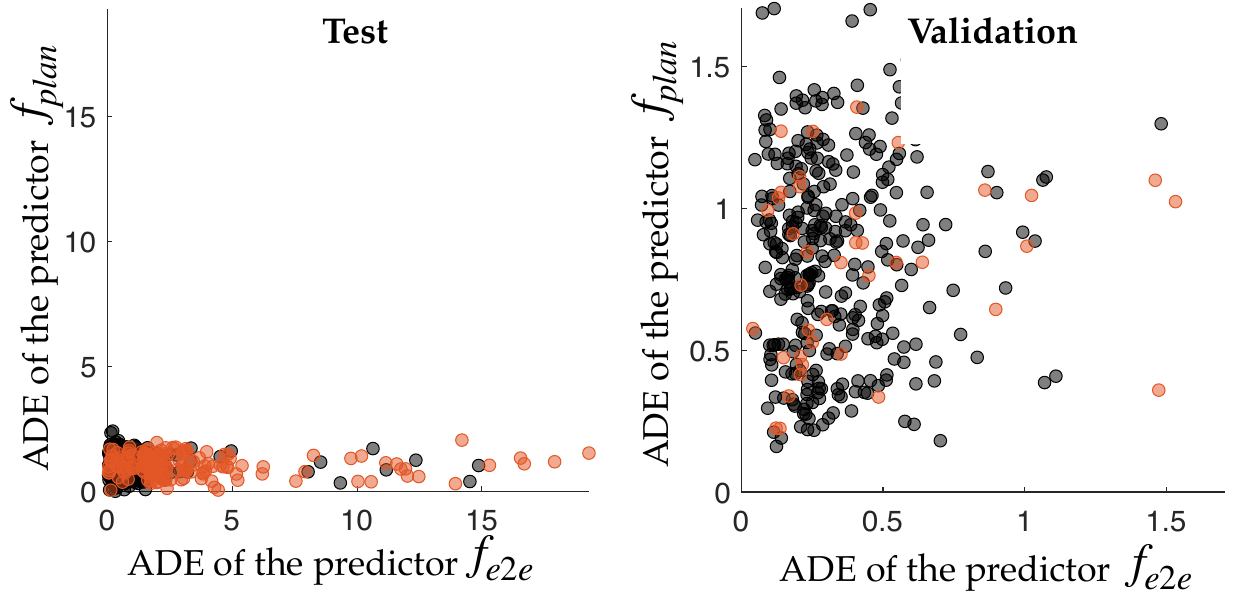}
	\caption{The ADE scatter with an ensemble metric as a performance detector (black: good cases for the lstm model; red: good cases for the irl model)}
	\label{fig:scatterwithensemble_exp2}
\end{figure}

\begin{figure}
	\centering
	\includegraphics[width=0.9\linewidth]{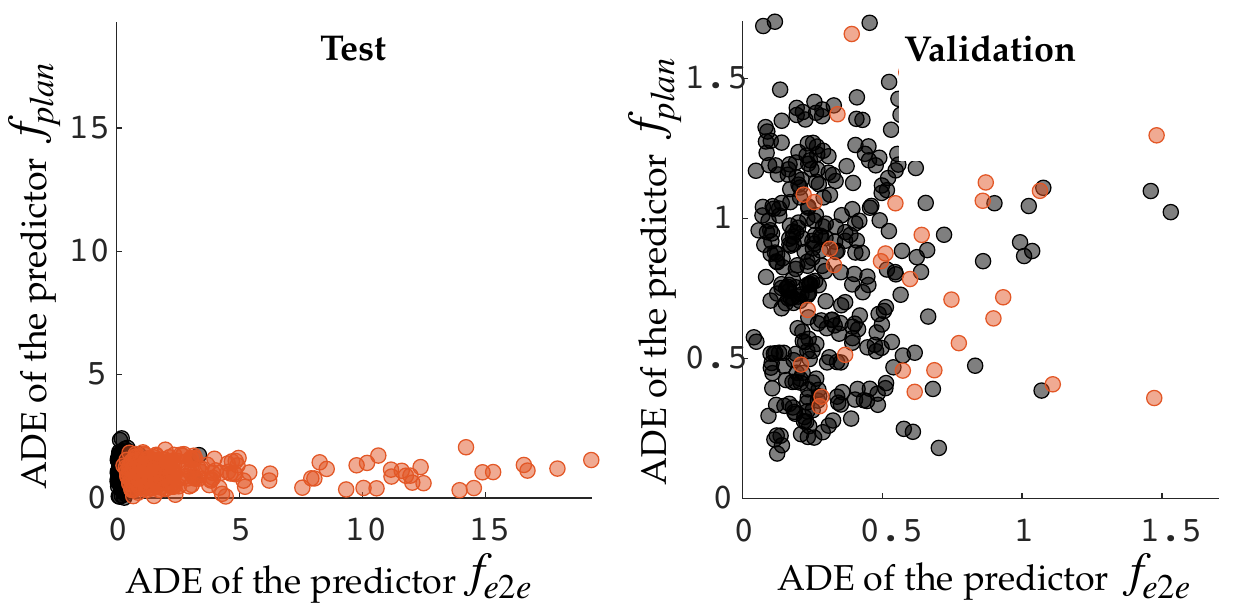}
	\caption{The ADE scatter with a trained classification model as a performance detector (black: good cases for the lstm model; red: good cases for the irl model)}
	\label{fig:scatterwithclassification_exp2}
\end{figure}

\begin{figure}
	\centering
	\includegraphics[width=0.9\linewidth]{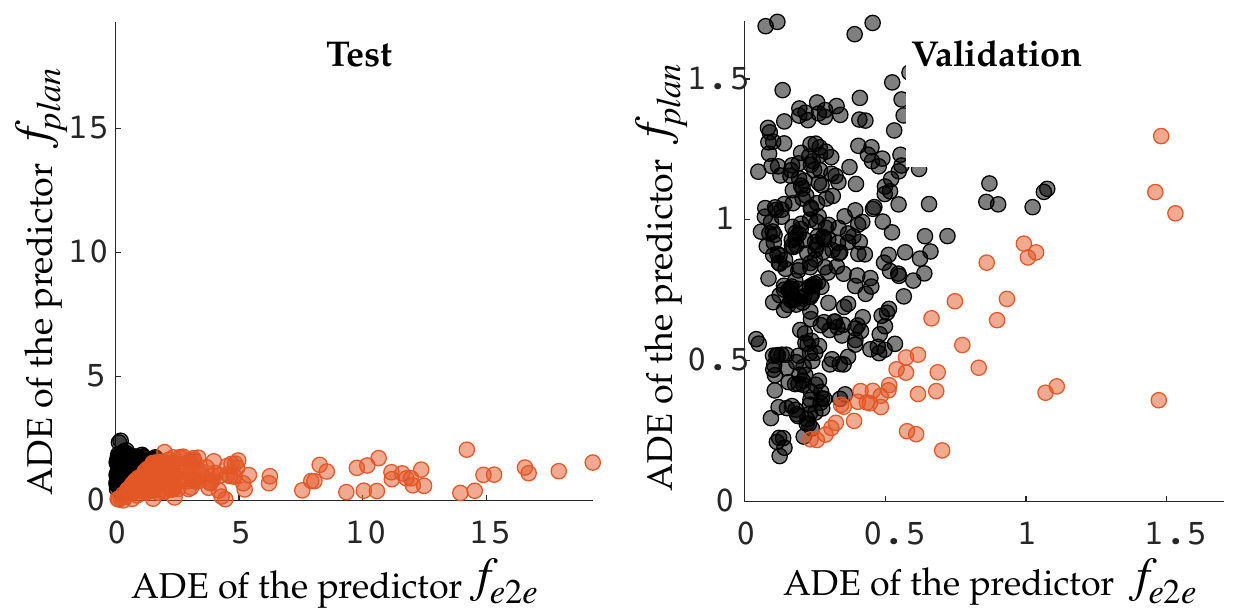}
	\caption{The ADE scatter with online Bayesian based on 30-step trajectories (black: good cases for the lstm model; red: good cases for the irl model)}
	\label{fig:scatterwithbayesian_exp2}
\end{figure}

\begin{figure}
	\centering
	\includegraphics[width=0.9\linewidth]{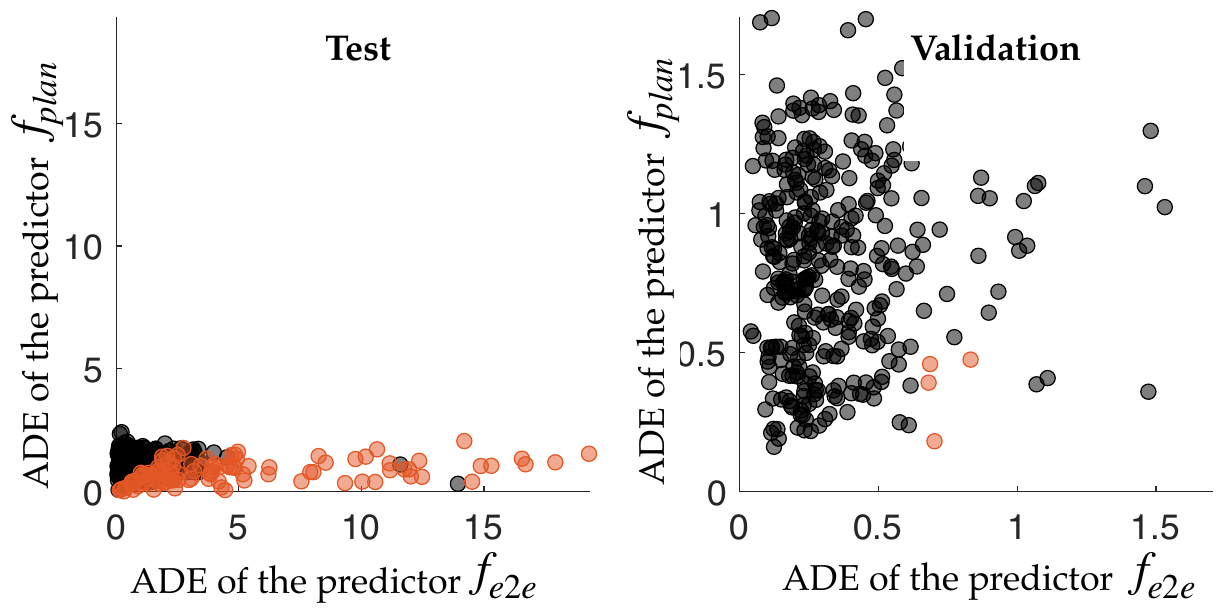}
	\caption{The ADE scatter with online Bayesian based on 5-step trajectories (black: good cases for the lstm model; red: good cases for the irl model)}
	\label{fig:scatterwith5stepbayesian_exp2}
\end{figure}

\figref{fig:scatterwithensemble_exp2} - \figref{fig:scatterwith5stepbayesian_exp2} show the scatter plots with different switching methods, again mirroring findings from Experiment I.


\begin{figure}[ht!]
	\begin{centering}
		\includegraphics[width=0.9\linewidth]{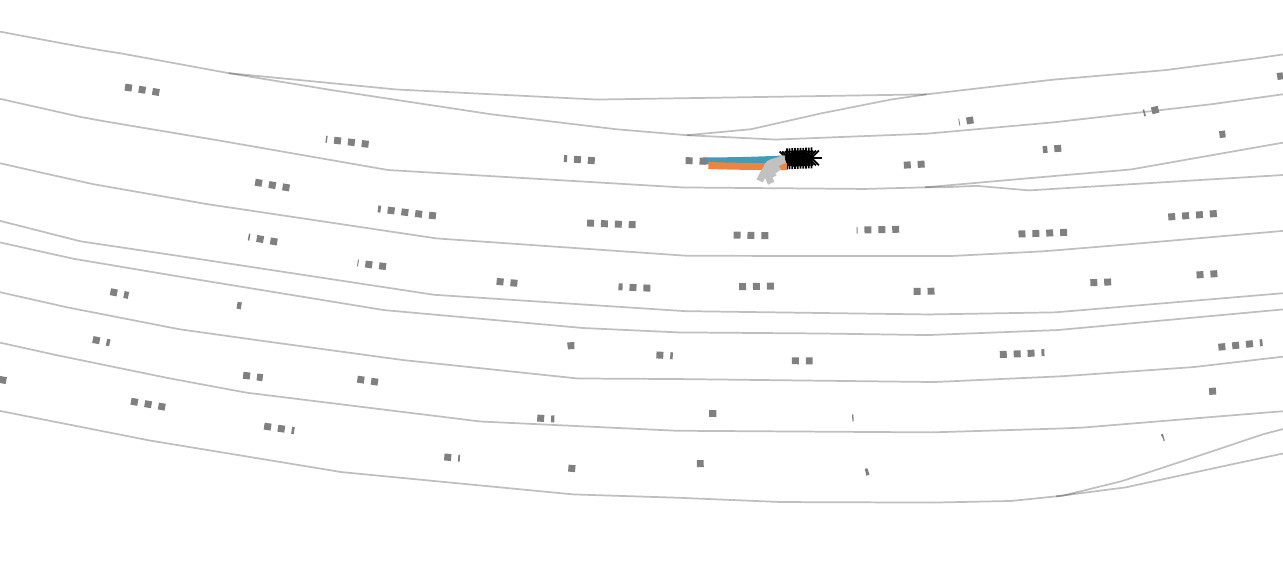}
		\includegraphics[width=0.9\linewidth]{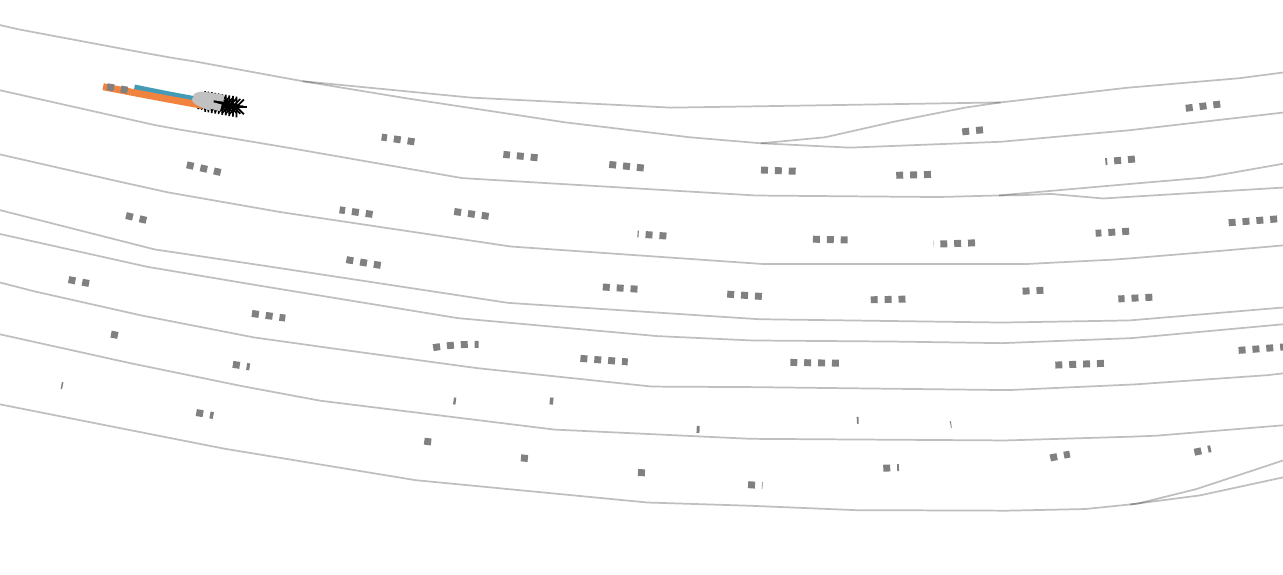}
	\end{centering}
	\caption{Failures of $f_{e2e}$ when tested on a new map. Its predictions in gray, $f_{plan}$ predictions in orange, and ground truth in blue.}
	\label{fig:exp2_failure_lstm_test_2}
\end{figure}

\figref{fig:exp2_failure_lstm_test_2} shows failures of the end-to-end predictor on an additional test map. 

\subsubsection{Experiment III}
\begin{figure}
    \centering
    \includegraphics[width=\columnwidth]{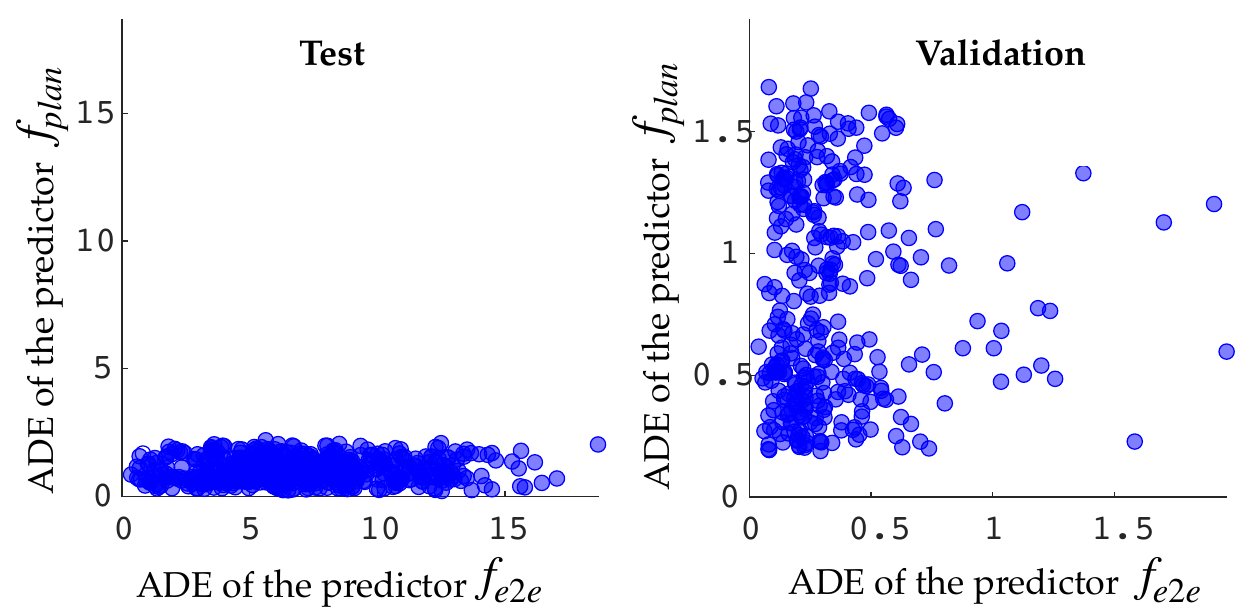}
    \caption{The performance of the two predictors in validation and test domains in Experiment III.}
    \label{fig:exp3_scatter}
\end{figure}

\figref{fig:exp3_scatter} shows the performance of the two predictors in validation and test under added noise.

\begin{figure}
	\centering
	\includegraphics[width=0.9\linewidth]{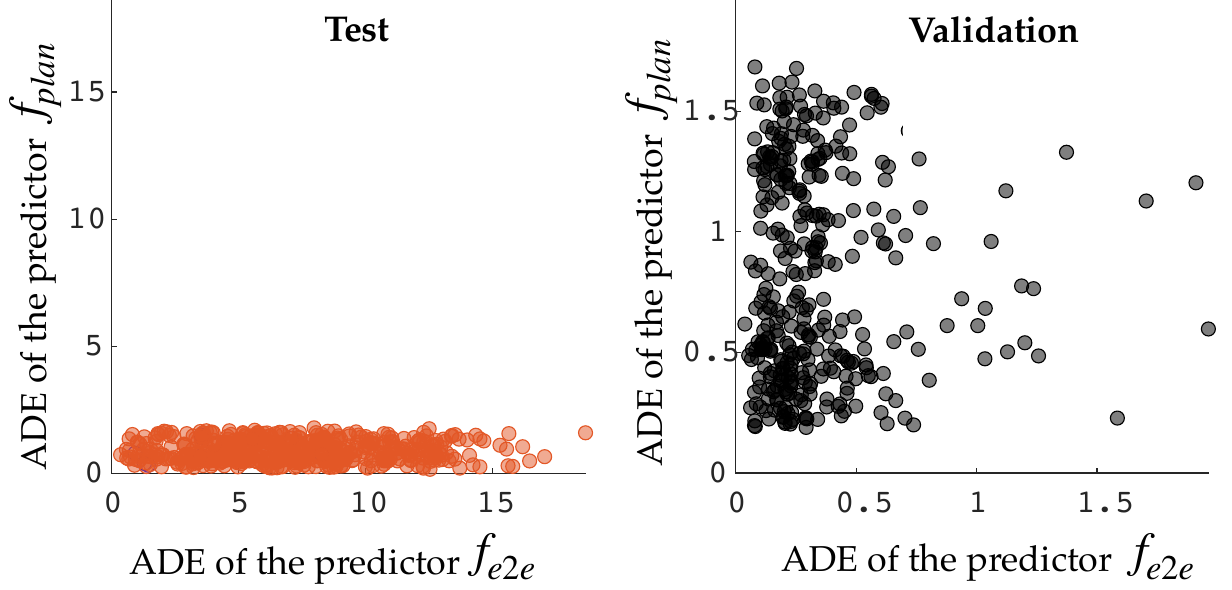}
	\caption{The ADE scatter with GAN as a scenario detector (black: good cases for the lstm model; red: good cases for the irl model)}
	\label{fig:scatterwithgan_exp3}
\end{figure}

\begin{figure}
	\centering
	\includegraphics[width=0.9\linewidth]{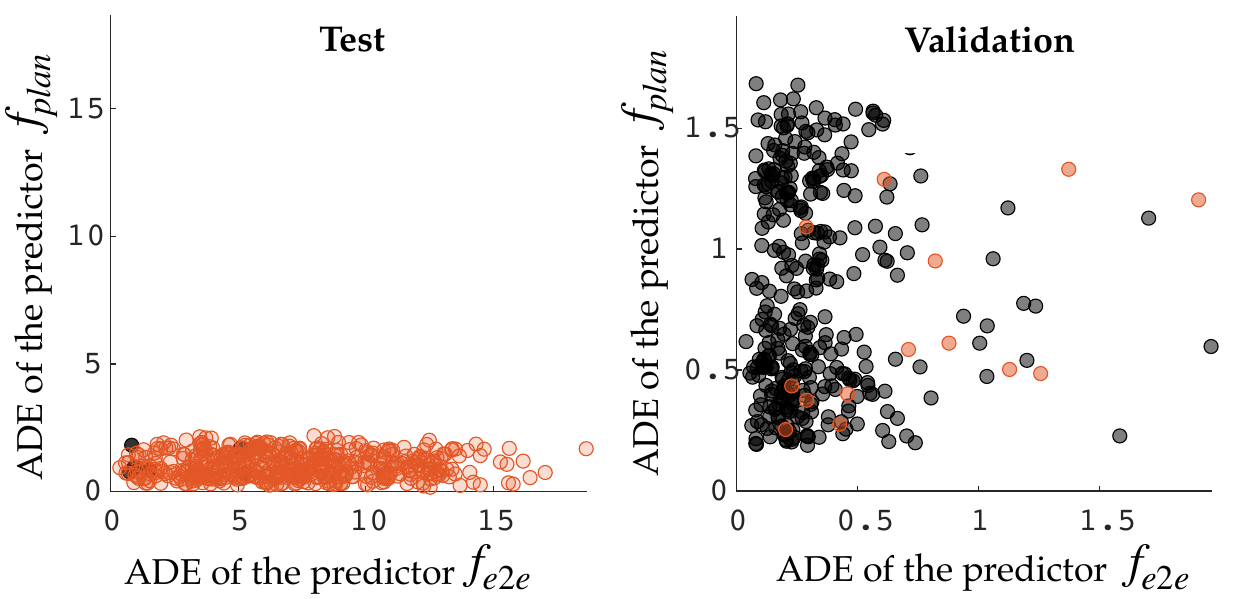}
	\caption{The ADE scatter with a trained classification model as a performance detector (black: good cases for the lstm model; red: good cases for the irl model)}
	\label{fig:scatterwithclassification_exp3}
\end{figure}

\begin{figure}
	\centering
	\includegraphics[width=0.9\linewidth]{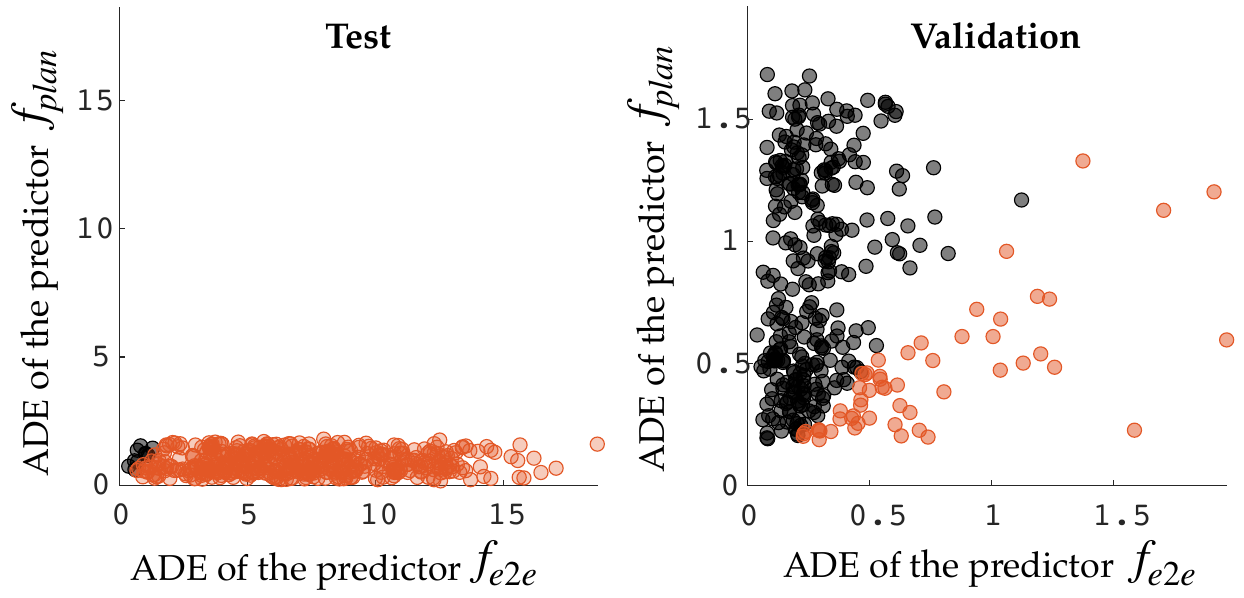}
	\caption{The ADE scatter with online bayesian based on 30-step trajectories (black: good cases for the lstm model; red: good cases for the irl model)}
	\label{fig:scatterwithfulltrajbayesian_exp3}
\end{figure}

\begin{figure}
	\centering
	\includegraphics[width=0.9\linewidth]{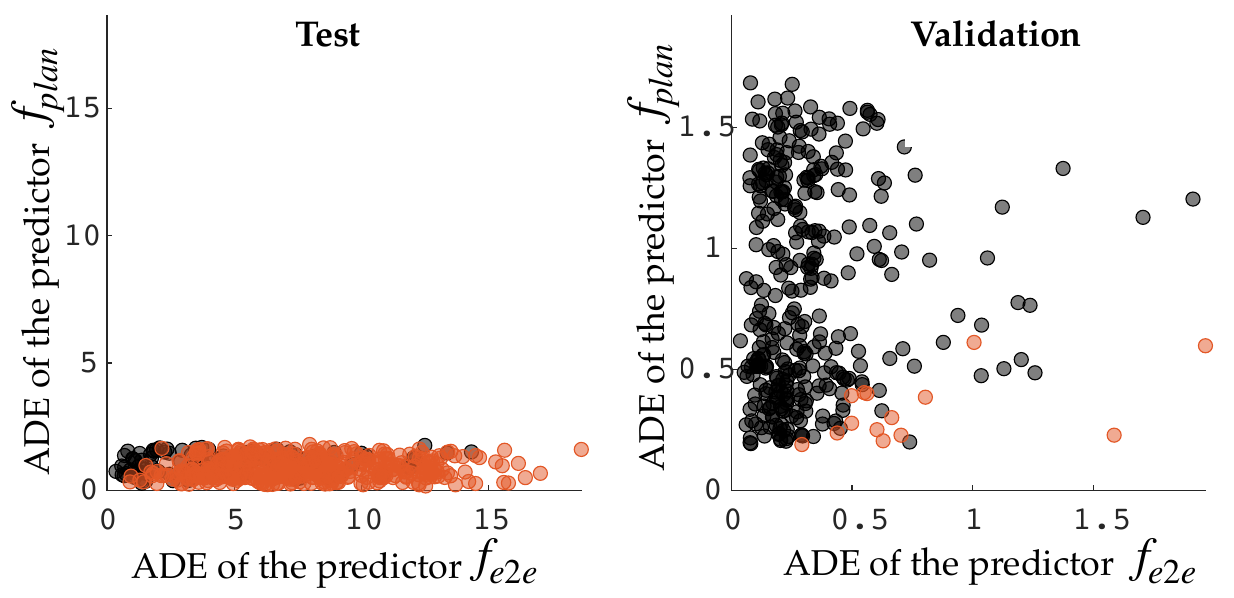}
	\caption{The ADE scatter with online bayesian based on 5-step trajectories (black: good cases for the lstm model; red: good cases for the irl model)}
	\label{fig:scatterwith5steptrajbayesian_exp3}
\end{figure}

\figref{fig:scatterwithgan_exp3} - \figref{fig:scatterwith5steptrajbayesian_exp3} show the scatter plots with different switching methods, again mirroring findings from Experiment I and II.

\begin{figure}
	\centering
	\includegraphics[width=0.9\linewidth]{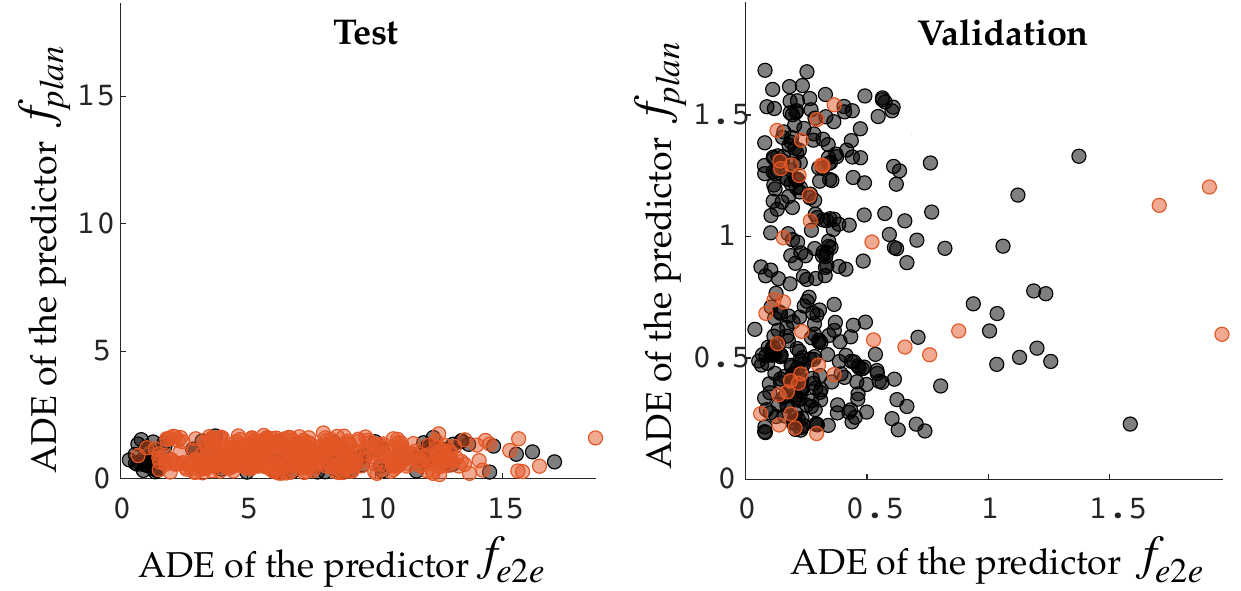}
	\caption{The ADE scatter with ensembling as a performance detector (black: good cases for the lstm model; red: good cases for the irl model)}
	\label{fig:scatterwithensemble_exp3}
\end{figure}
\end{document}